\documentclass[10pt,journal,compsoc]{IEEEtran}

\ifCLASSOPTIONcompsoc
  \usepackage[nocompress,noadjust]{cite}
\else
  \usepackage{cite}
\fi

\usepackage{hyperref}
\usepackage{enumitem}
\usepackage{booktabs}
\usepackage{times}
\usepackage{epsfig}
\usepackage{graphicx}
\usepackage{subfigure}
\usepackage{amsmath}
\usepackage{amssymb}
\usepackage{verbatim}
\usepackage{color}
\usepackage{colortbl}
\usepackage{xcolor}
\usepackage{ragged2e}
\usepackage{multicol}
\usepackage{gensymb}
\usepackage{tabularx}
\usepackage{multirow}
\usepackage{url}
\usepackage{xspace}
\usepackage{overpic}
\usepackage{wrapfig}
\usepackage[normalem]{ulem}

\newcommand{\etal}{\textit{et al.}}

\graphicspath{{./images/}}

\newcommand{\name}{iPUNet\xspace}

\newcommand{\Revised}[1]{{\color{black} #1}}

\begin{document}
\title{\name: Iterative Cross Field Guided Point Cloud Upsampling}
\author{Guangshun Wei,~
        Hao Pan,~
        Shaojie Zhuang,~
        Yuanfeng Zhou*,~
        Changjian Li

\IEEEcompsocitemizethanks{
\IEEEcompsocthanksitem Guangshun Wei, Shaojie Zhuang, and Yuanfeng Zhou are with the School of Software, Shandong University, Jinan 250101, P.R.China. E-mail: guangshunwei@gmail.com, sjzhuang@mail.sdu.edu.cn, yfzhou@sdu.edu.cn \protect
\IEEEcompsocthanksitem Hao Pan is with Microsoft Research Asia, Beijing, 100080, P.R.China. E-mail: haopan@microsoft.com\protect
\IEEEcompsocthanksitem Changjian Li is with the School of Informatics, University of Edinburgh, Edinburgh, EH8 9AB, United Kingdom. E-mail:Changjian.li@ed.ac.uk \protect
}

\thanks
{*Corresponding authors. Preprint Version.}
% {Manuscript received December 1, 2022; revised May 1, 2022.}
}

\markboth{IEEE TRANSACTIONS ON VISUALIZATION AND COMPUTER GRAPHICS, ~Vol.~*, No.~*, October~2023}%
{Shell \MakeLowercase{\textit{et al.}}: Bare Demo of IEEEtran.cls for Computer Society Journals}

\IEEEtitleabstractindextext{%

\begin{abstract}
Point clouds acquired by 3D scanning devices are often sparse, noisy, and non-uniform, causing a loss of geometric features. 
To facilitate the usability of point clouds in downstream applications, given such input, we present a learning-based point upsampling method, i.e., {\em \name}, which generates dense and uniform points at arbitrary ratios and better captures sharp features.
To generate feature-aware points, we introduce cross fields that are aligned to sharp geometric features by self-supervision to guide point generation.
Given cross field defined frames, we enable arbitrary ratio upsampling by learning at each input point a local parameterized surface.
The learned surface consumes the neighboring points and 2D tangent plane coordinates as input, and maps onto a continuous surface in 3D where arbitrary ratios of output points can be sampled.
To solve the non-uniformity of input points, on top of the cross field guided upsampling, we further introduce an iterative strategy that refines the point distribution by moving sparse points onto the desired continuous 3D surface in each iteration.
Within only a few iterations, the sparse points are evenly distributed and their corresponding dense samples are more uniform and better capture geometric features.
Through extensive evaluations on diverse scans of objects and scenes, we demonstrate that {\em \name} is robust to handle noisy and non-uniformly distributed inputs, and outperforms state-of-the-art point cloud upsampling methods.
\end{abstract}

}

\maketitle
\IEEEdisplaynontitleabstractindextext

\IEEEpeerreviewmaketitle

% ---------main text-------------
\IEEEraisesectionheading{\section{Introduction}}

\IEEEPARstart{P}{}oint cloud is one of the most popular 3D representations due to its flexibility, and is widely used in many applications, e.g., autonomous driving, robotics, rendering and modeling, etc. 
However, due to the limited resolution, point clouds acquired by consumption-level 3D scanning devices are often sparse, noisy, and non-uniform, causing a loss of geometric features. 
These imperfections greatly hinder the usability of point clouds in downstream applications, which poses the challenging task of point cloud {\em upsampling}, i.e., taking as input the imperfect data and producing a dense and uniform point cloud, while capturing sharp features as much as possible.

Various point cloud upsampling approaches have been proposed. Conventional optimization-based methods (\cite{Alexa2003Computing,Lipman2007,wu2015}) make use of shape priors like local smoothness. 
However, these methods are sensitive to noises and sparsity of the input, leading to poor robustness on complex shapes and large scenes. 
Along with the rapid development of neural networks, numerous data-driven methods with improved robustness have been proposed. 
Most of the upsampling neural networks run in a patch-based manner for generality and reduced computational cost (\cite{lipunet,wangpatch,fengneural}), where compact latent features are learned by considering the local geometry, and the features are then split and varied to generate new points. 
Moreover, enhancements have been developed, e.g. adversarial training for improving point uniformity \cite{lipunet} and edge-aware loss for sharp feature reconstruction \cite{yuecnet}. Although these data-driven approaches achieve state-of-the-art performance, they solve the challenges partially, e.g., \Revised{only Meta-PU \cite{9351772} and APU~\cite{dell2022arbitrary} support the dense sampling at arbitrary ratios, and }only EC-Net \cite{yuecnet} can enhance sharp features.

In this paper, we address the challenges in one framework by proposing {\em \name}, a learning-based point cloud upsampling method, which generates dense and uniform points at arbitrary ratios and better captures sharp features. 
Specifically, to enable feature aware upsampling with arbitrary ratios, we augment the local patches with feature-aligned cross fields.
The benefits of using cross field are two-fold. 
First, the cross field defines local frames for tangent spaces, on which we build parameterized surfaces embedded smoothly in 3D; the upsampling at arbitrary ratios can then be achieved trivially by sampling the tangent plane parameterization. 
Second, we align the cross field with geometric features through self-supervision, and the alignment naturally leads to sampling and generating feature-aligned dense points.
To deal with severe non-uniformity of input points, instead of trying to solve this problem in one pass, we propose an iterative strategy that refines the sparse input point distribution to obtain uniform upsampling. 
Specifically, in each iteration, we not only predict the dense upsampling, but also move the input sparse points toward better uniformity over the predicted surface; in the next iteration, the updated sparse points serve as the input again for cross field guided point upsampling. 
Within only a few iterations, the sparse input points are evenly distributed and their corresponding dense upsamples are more uniform and better capture geometric features.

We have conducted extensive experiments and comparisons to demonstrate the superior performance of \name against the state-of-the-art methods, as well as exhaustive ablation studies to validate our core technical designs. 
In particular, we have applied our method to real-world LiDAR scans and the results are remarkable, proving the robustness and applicability of our method in real-world applications.

In summary, the main contributions of this paper are as follows.
\begin{itemize}
    \item We introduce self-supervised cross field learning to enable feature-ware point upsampling at arbitrary ratios.
    \item We propose an iterative refinement strategy that greatly improves the uniformity of both the sparse input points and the upsampled dense points. 
    \item Combining the above two technical ingredients, we propose {\em \name}, a learning-based point cloud upsampling method that is robust to noise and sparsity of the input, generating dense and uniform points at arbitrary ratios, and better capturing sharp features. Our method outperforms the current state-of-the-art approaches, as validated through extensive experiments and ablation studies.
\end{itemize}

%\vspace{-0.3cm}
\section{Related work}

\noindent\textbf{Optimization-based upsampling.}
In order to generate dense points from the sparse input, optimization-based methods have been proposed to drive the point sampling based on global or local shape priors. 
Alexa \etal~\cite{Alexa2003Computing} proposed to sample vertices of Voronoi diagrams formed by neighboring points projected onto the local tangent plane. However, this method is sensitive to both noises and outliers. 
Subsequently, Lipman \etal~\cite{Lipman2007} developed a parametrization-free method that uses the locally optimal projection operator (LOP) for point resampling and surface reconstruction. 
Weighted LOP~\cite{Huang2009} was proposed to generate denoised and dense point clouds. 
Huang \etal~\cite{Huang2013} proposed an edge-aware resampling (EAR) algorithm to effectively preserve sharp features. It works in two passes: the areas far away from sharp edges are first upsampled, and then new points are progressively added toward the singular edges. 
Wu \etal~\cite{wu2015} proposed a point set consolidation method to fill large holes and complete missing regions by augmenting each surface point with an inner point that lies on the meso-skeleton. Notably, these methods rely heavily on explicit shape priors and demand careful finetuning of optimization parameters to work in different cases.
In comparison, our learning-based approach can be trained on diverse data distributions and perform robustly.

\vspace{1mm}
\noindent\textbf{Deep learning-based upsampling.}
With the rapid advances of point cloud-based neural networks, many methods have been designed to tackle various tasks, such as point cloud classification~\cite{Charles2017PointNet,xu2018spidercnn,wu2019pointconv}, segmentation~\cite{qi2017pointnet++,li2018pointcnn,wang2019dynamic}, denoising~\cite{RakotosaonaBGMO20,ChenWSXW20,zhang2020pointfilter}, and registration~\cite{fu2021robust}. For point cloud upsampling, quite a few approaches are proposed in recent years; we review the representative works in the following.

PU-Net~\cite{lipunet} presented the first data-driven point cloud upsampling method that runs in a patch-based manner. Its core idea is to learn multi-level features for each point and then split the expanded features to generate new dense points, while pushing them to distribute evenly around the desired surface. 
Then, EC-Net~\cite{yuecnet} was proposed to preserve sharp edges when upsampling the input, based on an edge-aware joint learning strategy. However, to make use of the edge prior, edge points need to be labeled manually, which is labor-intensive and time-consuming. 
Inspired by neural image super-resolution techniques, Wang \etal~\cite{wangpatch} designed a cascade of patch-based upsampling networks on different levels of details to achieve point cloud upsampling with high-fidelity geometric details.
PU-GAN~\cite{lipugan} uses the adversarial framework to synthesize uniformly distributed points, but its results are often noisy, especially around fine details. 
Qian \etal~\cite{qianpugeonet} proposed PUGeo-Net that incorporates discrete differential geometry into a point cloud upsampling network by learning the first and second fundamental forms. It can generate a relatively clean point cloud, but with many holes, if the input is sparse and non-uniform.

More recently, Meta-PU~\cite{9351772} was proposed as the first network that supports arbitrary ratio upsampling. However, their results are far from uniform and have local holes.
MAFU~\cite{9555219} designed a lightweight neural network for learning interpolation weights by analyzing the local geometry of the input point cloud, which can generate uniform points for the upsampling task.
METZER \etal~\cite{metzer2021self} designed a self-supervised upsampling method to obtain clean and outlier-free dense points through an iterative training strategy.
Dis-PU~\cite{lidisup} designed the point generator and the spatial refiner to first generate a dense point cloud and then further improve its quality.
Feng \etal~\cite{fengneural} constructed neural fields through the local isomorphism between the 2D parametric domain and the 3D local patch, and then integrated the local neural fields to form a global surface where new points can be sampled. 
PU-flow \cite{mao2022pu} exploited the invertible normalizing flows to transform points between Euclidean and latent spaces, which can produce dense points uniformly distributed on the underlying surface. \Revised{APU~\cite{dell2022arbitrary} mapped sparse inputs to a spherical mixed Gaussian distribution from which any number of points can be sampled. Zhao \etal~\cite{zhao2022self} proposed to define point cloud upsampling as the task of finding the nearest projection point on seed points, which allows for simultaneous self-supervised learning and flexible magnification in point cloud upsampling.}
All the recent methods improve the quality of the upsampled points significantly; however, they are sensitive to the sparsity and non-uniformity of input points, which often lead to many holes in the results (c.f. Sec.~\ref{sec:results}).

Our method builds upon a learned cross field to provide the basis for noise-free and feature-aware upsampling at arbitrary ratios. 
We further propose an iterative updating scheme that refines the sparse point distribution to achieve uniform upsampling. 
Comparisons with state-of-the-art methods both qualitatively and quantitatively demonstrate our superior performance (Sec.~\ref{sec:results}).

%%%%%%%%%%%%%Pipeline Figure%%%%%%%%%%%%%
\begin{figure*}[!t]
\centering
\begin{overpic}[width=\linewidth]{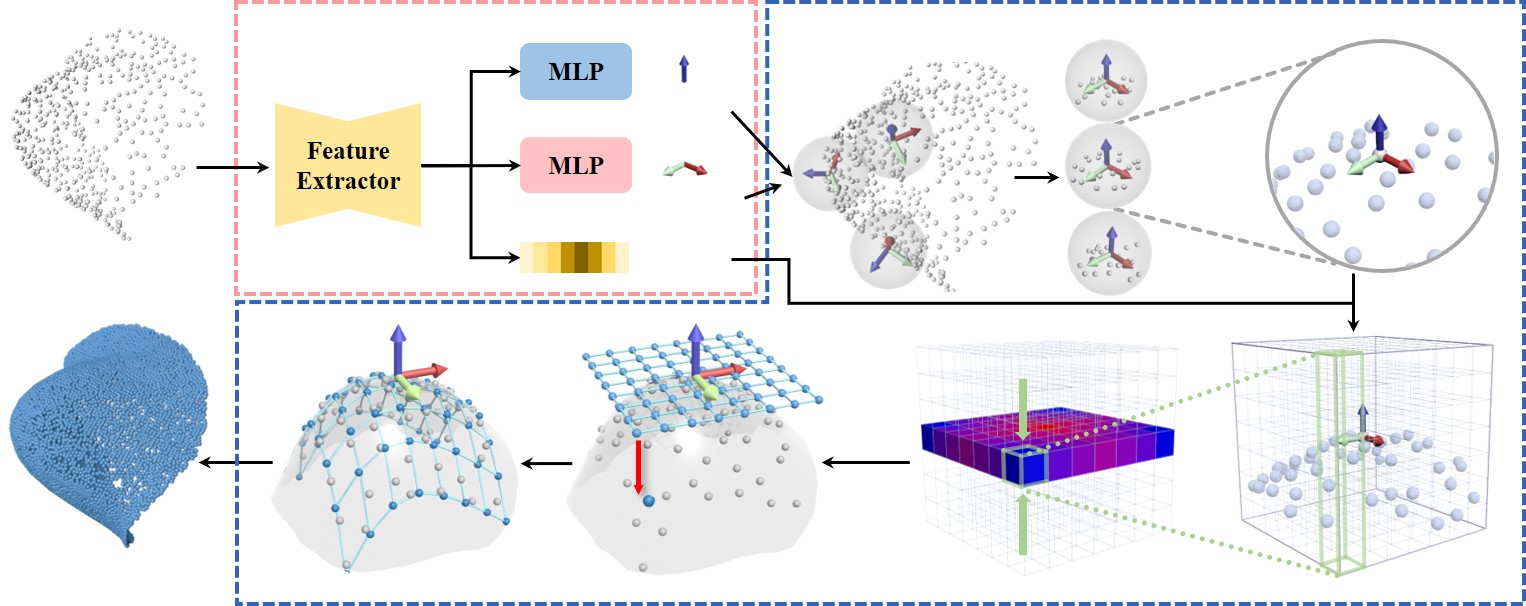} % grid,tics=2,
     \put(3.5, 22) {\small Input ($m\times$3)}
     \put(2.5, 1.5) {\small Output ($mr\times$3)}
     \put(16.2, 37.6) {\small \textbf{Field and normal estimation}}
     \put(19,23) {\small (per-point)}
     \put(41.6, 26) {\footnotesize Cross field}
     \put(42.2, 32) {\footnotesize Normal}
     \put(42.8, 22.5) {\footnotesize Feature}
     \put(51, 37.6) {\small \textbf{Mapping function learning}}
     \put(56, 32.3) {\small 1}
     \put(52, 30.3) {\small 2}
     \put(55, 24) {\small 3}
     \put(69, 36) {\small 1}
     \put(69, 30) {\small 2}
     \put(69, 24) {\small 3}
     \put(75, 36.5) {\footnotesize (a) Transformation}
     \put(86, 33) {\small $K_2$-neighbors }
     \put(89, 19.7) {\footnotesize (b) Voxelization}
     \put(83, 0.8) {\footnotesize (c) Feature inpainting}
     \put(61.5, 0.8) {\footnotesize (d) Feature compression}
     \put(40.5, 0.8) {\footnotesize (e) Offset prediction}
     \put(20.5,0.8) {\footnotesize (f) Point projection}
     \put(56.8, 10) {\small $\varphi$}
     \put(54.4, 8) {\footnotesize mapping}
     \put(54.6, 6.5) {\footnotesize function}
     \put(39.8,10.5) {\small $p_t^{i}$}
     \put(40.5,5.5) {\small $q_t^{i}$}
     \put(42.5, 8.5) {\small $o_t^{i}$}
    \end{overpic}
\caption{\textbf{Overview of our network}. Our {\em \name} has two core components, i.e., field and normal estimation and mapping function learning. In the first component, the per-point cross field, normal, and feature are estimated (Sec. \ref{subsec:field_estimation}), which form the basis for the later component. Then, in the second component, given cross field defined frames, we learn a mapping function that maps any point on the tangent plane to the target shape (Sec.~\ref{subsec:mapping_func}). The learned mapping function enables us to upsample at arbitrary ratios.}
\label{fig:pipeline}
\end{figure*}
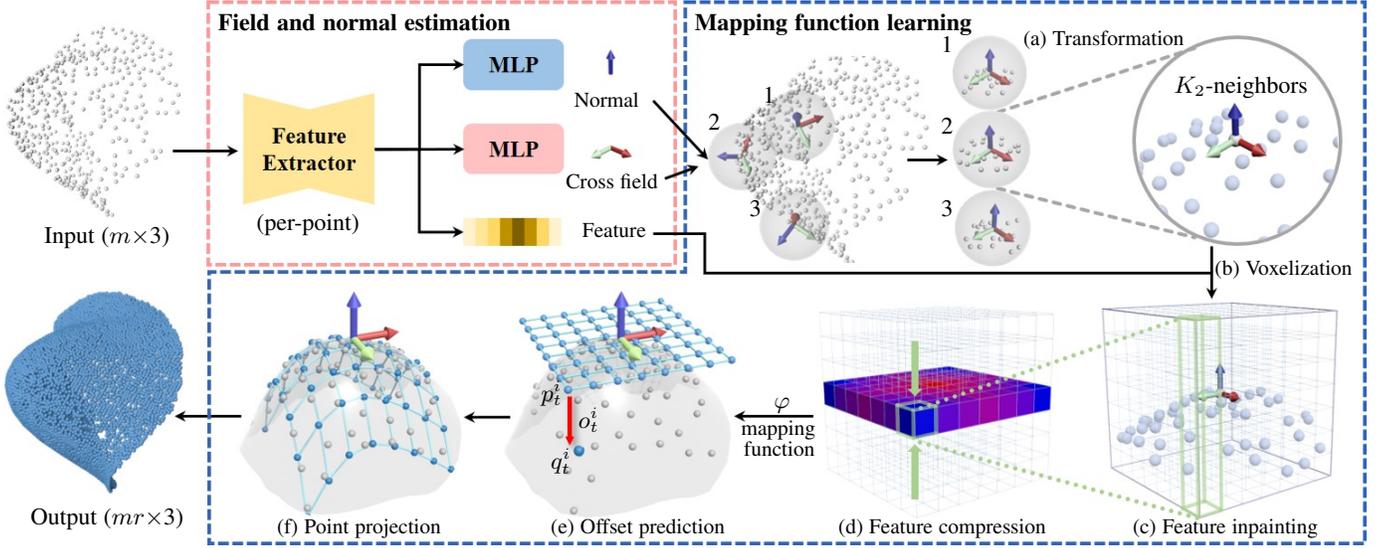

%%%%%%%%%%%%%%%%%%%%%%%%%%%%%%%%%%%%%%%

\vspace{1mm}
\noindent\textbf{Feature aligned cross field estimation.}
Cross fields (also called 4-rotational symmetric field/4-RoSy fields) defined on a surface have been introduced and thoroughly researched in Computer Graphics (\cite{hertzmann2000illustrating,lai2009metric,bommes2009mixed,diamanti2014designing}). 
Since a cross field aligned with sharp features captures the geometry of the surface, it is widely used in many applications - NPR rendering~\cite{hertzmann2000illustrating}, surface quadrangulation~\cite{JakobTPS15}, and parameterization~\cite{SorkineHornung14a}, to name a few. 
In recent years, \cite{2019TextureNet,yang2020pfcnn} utilized cross fields to define convolution neural networks on a surface. 
Huanng \etal~\cite{DBLPHuang2019} proposed to predict dense cross fields as canonical 3D coordinate frames from a single RGB image. 
In this paper, we learn cross fields aligned with sharp features via self-supervised learning, and use them to achieve feature-aware arbitrary ratio upsampling.

%\vspace{-0.2cm}
\section{Method}
\label{sec:method}

\subsection{Overview}
\label{subsec:overview}

Given a sparse, noisy, and non-uniform point cloud $X=\{ p_i  \}_{i=1}^m$ and an \emph{arbitrary} upsampling ratio $r$ ($r \in R$, and $r>1.0$), we aim to generate a dense and uniform point cloud $Y=\{ p_j \}_{j=1}^{M}$ ($M=\lfloor r*m \rfloor$) that faithfully captures the underlying geometry with better sharp feature alignment. To this end, we have proposed \emph{\name}, a novel point cloud upsampling method that builds upon the point-based cross field. As benefiting from the learned cross-field, our method can take full consideration of the underlying geometry instead of paying more attention to point positions solely as in existing works. 

Fig.~\ref{fig:pipeline} demonstrates the overview of our method, which consists of two main components. Specifically, we first learn the point-wise cross-field and normal, so that we can construct the point-wise parameterized surface (i.e., the tangent plane) as well as the local coordinate. Then, for each point, we transform its neighbors within a certain distance to the corresponding coordinate and learn the mapping of arbitrary surface points to the target 3D surface by predicting offset vectors.
In practice, the mapping is learned in a discrete manner, i.e., 
given any input point along with the learned local \Revised{frame}, we first discretize the 3D space centered at that point with a 3D grid, where we learn the geometric features for each grid point with the help of the transformed neighbors. After feature learning, we collapse the $3d$ grid to its $2d$ counterpart (i.e., coordinates of the tangent plane of the center point) by feature resizing, and finally predict the offset vectors (i.e., the distance and direction) for each $2d$ grid point to the target surface.

In addition to the cross-field guided upsampling, we further propose an iterative strategy to learn the mapping and update positions of input sparse points iteratively, which significantly improves the uniformity of both the up-sampled and input point clouds, as well as the robustness of the proposed algorithm dealing with noises and sparsity. 
In the following, we elaborate on the technical details.

%\vspace{-0.2cm}
\subsection{Cross Field and Normal Estimation}
\label{subsec:field_estimation}
Inspired by the surface-based cross-field, we introduce and define the point-based cross-field in the following. Given a point $p_i$, the cross field lies in its tangent plane defined by its normal $n_i \in \mathbb{R}^3$ (i.e. orthogonality), while the cross field itself is four-rotationally symmetric (4-RoSy) and can be represented by a unit vector $\theta_i \in \mathbb{R}^3$ up to a rotation by $k\pi/2$, where $k \in \{0,1,2,3\}$.

To estimate the normal and cross field for the input point cloud $X$, as shown in Fig.~\ref{fig:pipeline}, we employ the DGCNN~\cite{wang2019dynamic} as our feature extractor to obtain the latent feature $F \in \mathbb{R}^{m \times 128}$, that is further sent to a few MLP layers to derive our per-point cross fields $\{ \widetilde{\theta}_i\}_1^m$, normals $\{ \widetilde{n}_i\}_1^m$ and features $\{ f_i\}_1^m$ ($f_i \in \mathbb{R}^{128}$).

\vspace{1mm}
\textbf{Training Losses.} We supervise the normal learning by minimizing the cosine similarity between the predicted ($\widetilde{n}_i$) and the corresponding ground truth ($n^*_i$) normals:
\begin{equation}
    L_{normal}=\sum_{i=1}^{m} \left( 1- |\frac{ \langle \widetilde{n}_i, n^*_i \rangle}{\| \widetilde{n}_i\| \| n^*_i\|} | \right).
\end{equation}
As for the cross field, it is orthogonal to the corresponding normal by design, which can be described by the loss term:
\begin{equation}
	L_{field\_normal}=\sum_{i=1}^{m}  | \langle \widetilde{n}_i, \widetilde{\theta}_i \rangle |.
\end{equation}
We usually solve the cross field via mutual smoothness, while the definition of smoothness differs. Since we want to obtain sharp feature-aligned cross fields, such that the upsampled points are feature-aware. Thus, inspired by \cite{JakobTPS15}, we combine both smoothness and alignment requirements into one loss term, as follows:
%\vspace{-0.2cm}
\begin{equation}
\label{Eq:fieldsmooth}
	L_{field\_smooth}= \sum_{i=1}^{m} \sum_{i' \in K_1(i)} L_{s} \left( \widetilde{\theta}_i, \widetilde{\theta}_{i'} \right),
\end{equation}
where $K_1(i)$ is the index set of $K_1$-nearest neighbors of point $p_i$, while $L_{s}$ is defined as:
\begin{equation}
\begin{array}{l}
	L_{s} \left( \widetilde{\theta}_i, \widetilde{\theta}_{i'} \right)= min\left( 1-|
	\frac{ \langle \widetilde{\theta}_i, \widetilde{\theta}_{i'} \rangle}{\| \widetilde{\theta}_i\| \| \widetilde{\theta}_{i'}\|} |,
	1-|\frac{ \langle rot \left( \widetilde{\theta}_i \right),  \widetilde{\theta}_{i'} \rangle}{\| rot \left(\widetilde{\theta}_i \right)\| \|  \widetilde{\theta}_{i'}\|}|
	\right),
	\end{array}
\end{equation}
where $rot \left(\widetilde{\theta}_i \right)$ means to rotate $\widetilde{\theta}_i$ by $k\pi/2$ on the tangent plane, \Revised{and the operation $\langle a, b\rangle$ is the dot product}.

\setlength{\columnsep}{0.05in}
\begin{wrapfigure}[6]{r}{0.35\linewidth}
    \begin{overpic}[width=0.9\linewidth]{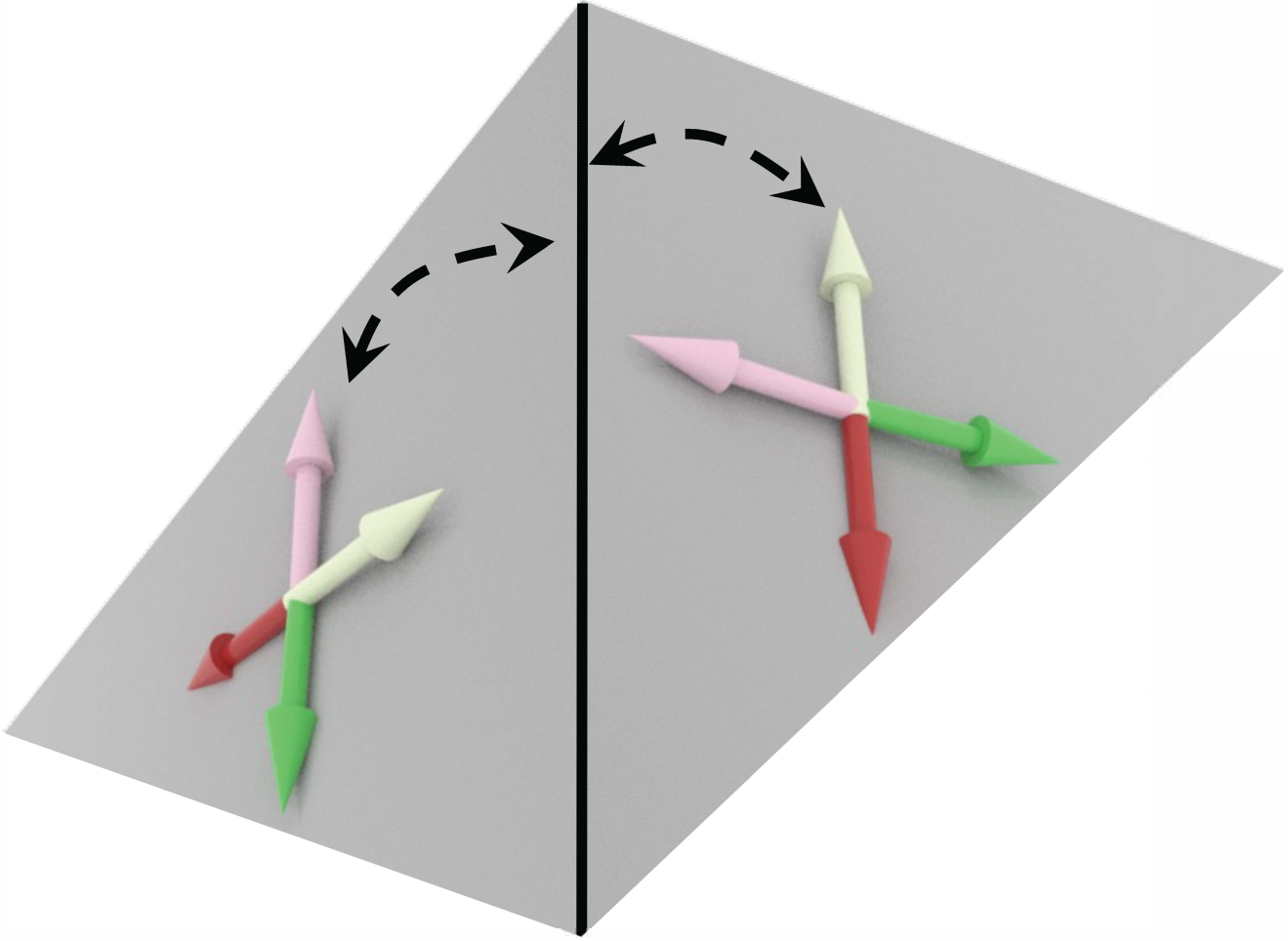} 
    \put(11,36) {\small $\theta_i$}
    \put(52,32) {\small $\theta_j$}
    \end{overpic}
\end{wrapfigure}
The rationale behind this is that when measuring the smoothness of two fields crossing a sharp edge, the loss obtains the minimum value only if they both align with the common edge (see inset). By using this self-supervised learning strategy, the learned fields are naturally aligned with sharp features achieving our goal.

\subsection{Mapping Function Learning}
\label{subsec:mapping_func}

Having the learned point-wise cross field and the normal, we can derive the parametric surface for any input point. The goal of this component is to learn a mapping function that maps points on the learned surface to the target continuous 3D surface, where we can sample new points. We achieve this goal in a discrete manner.

Specifically, given a input point $p_i$, we have obtained the local frames represented by $\left(\widetilde{\theta}_i, \widetilde{n}_i \times \widetilde{\theta}_i, \widetilde{n}_i \right)$, and the learned feature vector $f_i$. Firstly, we select the $K_2$-nearest neighbors within a radius \Revised{$r_n$} as the most related points to represent the local geometry (note, $K_2(i)$ denotes the index set of those selected points) and transform those points into the space of $p_i$ via a simple rotational transformation:
\begin{equation}
    p_j^i = (p_j - p_i) * R_i, \quad j \in K_2(i),
    \label{eq:transform}
\end{equation}
where the rotation matrix $R_i =  \begin{bmatrix} \widetilde{\theta}_i, \widetilde{n}_i \times \widetilde{\theta}_i, \widetilde{n}_i \end{bmatrix}$, as shown in Fig.~\ref{fig:pipeline}(a).
Secondly, we discretize the 3D space around $p_i$ according to its frames with voxel grids that are centered at $p_i$ with dimension $d \times d \times d$ and spatial spacing of $2*\Revised{r_n}/d$, as shown in Fig.~\ref{fig:pipeline}(b). Note here, the $z$=0 level set is the tangent plane (the learned parametric plane).
Thirdly, having the grid, we find the nearest voxel for each transformed point $p_j^i$, and simultaneously copy the feature $f_j$ \Revised{(128-dim)} into the corresponding voxel. In the meanwhile, for voxels without any existing transformed points, we assign $\textbf{0}$ vectors \Revised{(128-dim)} instead. We then employ a series of 3D convolutions as our feature inpaintor that takes as input the initial voxel features $F_v^{d\times d\times d \times 128}$ to learn a new feature $F_{vnew}^{d\times d \times d \times c}$ (see Fig.~\ref{fig:pipeline}(c)). 

% mapping function
The goal of our mapping function is to map each point on the parametric surface to the target 3D surface, where we can sample new points. To this end, given $p_i$, we first define the grid points on the tangent plane as $\{p_t^i \in \mathbb{R}^3 \; | \; t \in [0, d\times d), \; p_t^i(z)=0 \}$. In order to learn the mapping function, we compress the 3D grid space along the z-axis (i.e., the normal direction) into the $z=0$ level set by simply reshaping the feature $F_{vnew}^{d\times d \times d \times c}$ into $F_t^{d\times d \times (d \times c)}$. \Revised{We then use a series of 2D convolutions to further extract the information} such that each $p_t^i$ will have a feature vector $f_t^i \in \mathbb{R}^{d\times c}$ (see Fig.~\ref{fig:pipeline}(d)).

We implement the neural mapping function $\varphi$ via a \Revised{point-wise multilayer perceptron (MLP)} producing the offset vectors (as shown in Fig~\ref{fig:pipeline}(e)):
\begin{equation}
o_t^i = \varphi(p_t^i, f_t^i), \; o_t^i \in \mathbb{R}^3
\end{equation}
and we simply add the offset to the grid point to obtain the 3D point on the target 3D surface (as presented in Fig~\ref{fig:pipeline}(f)):
\begin{equation}
    q_{t}^{i} = p_t^i + o_t^i.
\end{equation}

% point cloud upsampling
\textbf{Upsampling.}
With the mapping function, we can sample arbitrary points on the tangent plane and map them to the desired 3D surface to derive the resulting 3D points. However, the 3D point is in the local coordinate of input $p_i$. Thus, we first transform the points back to the input space by a simple inverse transformation of Eq.~\ref{eq:transform}:
\begin{equation}
    \widetilde{p}_{t}^i = q_{t}^i * R^{-1} + p_i.
\end{equation}
Then, we collect all the transformed points from all input points into the resulting point set $\widetilde{Y} = \{ \widetilde{p}_{t}^i \in \mathbb{R}^3 \; | \; t \in [0, d\times d), \; i \in [1, m] \}$. And we supervise the training by measuring the Chamfer distance between the predicted ($\widetilde{Y}$) and the ground truth ($Y^*$) point cloud:
\begin{equation}
    L_{CD}(Y^*,\widetilde{Y})=\mathop {\sum}\limits_{y \in Y^*} \mathop {\min }\limits_{y' \in \widetilde{Y}} \| y-y' \|_2^2 +\mathop {\sum}\limits_{y' \in \widetilde{Y}} \mathop {\min }\limits_{y \in Y^*} \| y'-y \|_2^2.
\end{equation}
Note here, the upsampling ratio $r$ does not matter the training, we care more about the field and mapping function learning. Thus, for efficiency purposes, in each \emph{training} iteration, we randomly select eight grid points from the tangent grid of each input point. The randomness of the selection and the large iteration number ensure that all the grid points are involved in the mapping function learning, leading to a quick and sufficient training procedure. 

Till now, we can train the whole upsampling pipeline using the training loss:
\begin{equation}
\begin{array}{cc}
     L_{one\_pass}(Y^*, \widetilde{Y})=L_{normal}+\lambda_0 L_{field\_normal}&  \\\\
      +L_{field\_smooth}+\lambda_1L_{CD}(Y^*, \widetilde{Y}).
\end{array}
\label{eq:onepass}
\end{equation}

% Arbitrary Sampling.
\Revised{
\textbf{Arbitrary Upsampling at Testing Time.}
Although in the training process, we only sample some of the grid points for efficient mapping function learning, the learned mapping function is sufficient to transform arbitrary points within the tangent plane to the desired surface given the point position and the feature. Thus, at testing time, when the user specified $r$ is less than the number of grid points (i.e., $d \times d$), we first consider all grid points and randomly select $r$ points from them. 
\setlength{\columnsep}{0.05in}
\begin{wrapfigure}[6]{r}{0.55\linewidth}
    \begin{overpic}[width=0.9\linewidth]{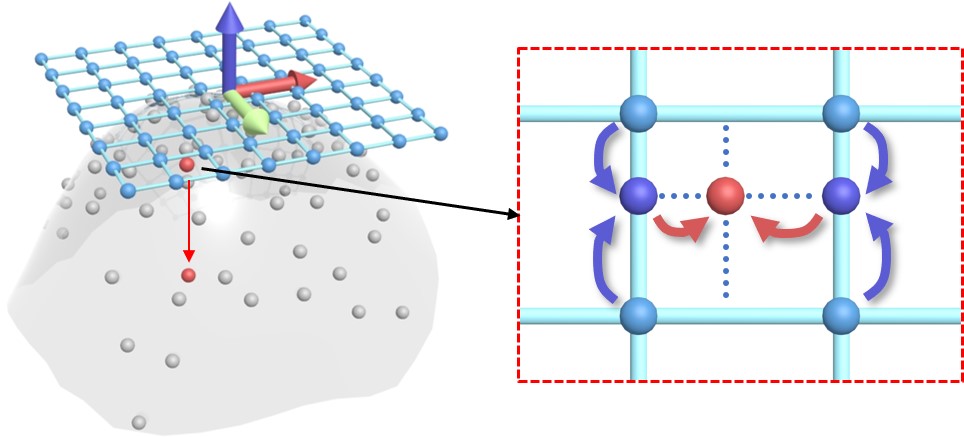}
    \end{overpic}
\end{wrapfigure}
However, when the specified ratio $r$ is over the number of grid points, we can randomly sample non-grid points in the tangent plane, and the corresponding features sent to the mapping function can be obtained via bilinear feature interpolation (see the inset figure). 

In Fig. \ref{fig:diff_ratio}, we have demonstrated upsampled points at arbitrary ratios.
}

\subsection{Iterative Update Strategy}
\label{subsec:iter_update}

%%%%%%%%%%%%%Iterative Updating%%%%%%%%%%%%%%%%%%
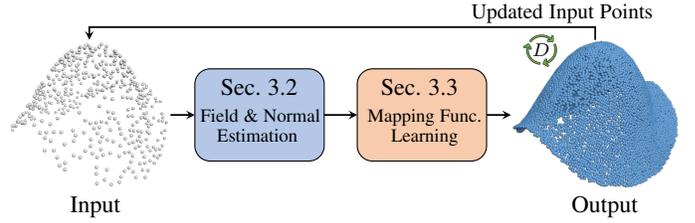
\begin{figure}[!t]
\centering
\begin{overpic}[width=\linewidth]{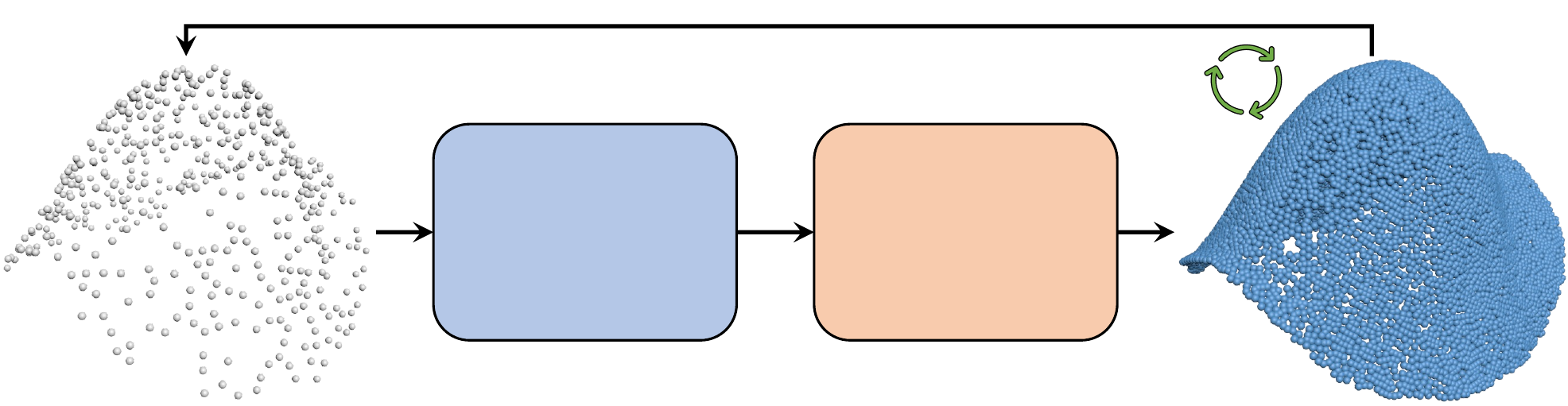}
    \put(9,-3.5) {\small Input}
    \put(84,-3.5) {\small Output}
    \put(69,25.5) {\footnotesize Updated Input Points}
    \put(78.1,19.8) {\scriptsize $D$}
    \put(31.5,14) {\small Sec.~\ref{subsec:field_estimation}}
    \put(28.5,10) {\scriptsize Field \& Normal}
    \put(31,7) {\scriptsize Estimation}
    \put(55.5,14) {\small Sec.~\ref{subsec:mapping_func}}
    \put(53.5,10) {\scriptsize Mapping Func.}
    \put(57,7) {\scriptsize Learning}
\end{overpic}
\caption{\textbf{Iterative Updating Strategy}. In the iterative updating strategy, after each iteration, we feed back the updated input points as the new input to our pipeline again, and we take the prediction from the last iteration as our final upsampled points.}
\label{fig:iter_strategy}
\end{figure}
%%%%%%%%%%%%%%%%%%%%%%%%%%%%%%%%%%%%%%%%%%%%%%%%

Considering that the input point cloud is sparse and non-uniform, the resulting upsampled points are often non-uniform and have holes. To address this challenge, we propose an iterative strategy, as shown in Fig.~\ref{fig:iter_strategy}. The idea is that in the upsampling process, we not only care about the quality of the new points but also drive the original input points to be distributed uniformly by enforcing them to move to the target 3D surface as well. As a result, the improved input points significantly benefit the field estimation and mapping function learning leading to more uniform upsampled points.

Specifically, for each training iteration, we further employ $D$ times inner iterations ($iter \in [0,D)$). Because, each input point is the center grid point in its tangent plane, and the offset from that point to the 3D surface is also calculated by the mapping function. Thus, along with the Chamfer distance between the upsampled (denoted as $\widetilde{Y}_{iter}$) and the ground truth points, we particularly minimize the Chamfer distance of the \emph{updated} input points (denoted as $\widetilde{X}_{iter}$) and the ground truth, as in the following:
\begin{equation}
    L_{uniform} = \lambda_{u}L_{CD}(Y^*, \widetilde{X}_{iter}),
\end{equation}
and the total loss for one iteration becomes:
\begin{equation}
    L_{total} = L_{one\_pass}(Y^*, \widetilde{Y}_{iter}) + L_{uniform}.
    \label{eq:total}
\end{equation}

Subsequently, the updated input points in the $i$-$th$ iteration will serve as the input to the ($i+$1)-$th$ iteration, and we repeat the aforementioned cross-field and normal estimation, and the mapping function learning processes until we reach $D$ total iterations. In the training stage, $D$ is set to be 10 for efficiency, while in the testing stage, $D$ can be set to different numbers by trading off efficiency and quality (10 is also used in the testing stage throughout this paper). As for the final result, we take the upsampled points in the last iteration, i.e., $\widetilde{Y}_{D-1}$, and a bonus is that we can simultaneously obtain more uniform input points, i.e., $\widetilde{X}_{D-1}$ (see examples in Sec.~\ref{sec:results}).

\subsection{Implementation and Training Details}
\textbf{Implementation.} 
We implement our framework using Pytorch, and the detailed network structure can be found in the supplemental material. In the following, we introduce some key parameters. When measuring the smoothness of the cross-field, we find the $K_1=6$ nearest neighbors, while we set $K_2=48$ and $\Revised{r_n}=1/3$ of the diagonal length of the input when cropping points to represent local geometry. In the mapping function learning, the dimension $d$ of the discretized 3D space is set to 7, and the feature dimension $c=64$. To balance loss terms in Eqs.~\ref{eq:onepass} and \ref{eq:total}, $\lambda_0$ is set to 0.1, $\lambda_1$ is set to 200, and $\lambda_u$ is set to 0.4.

\vspace{1mm}
\noindent\textbf{Network training.}
We train our whole network in an end-to-end manner for 200 epochs with a fixed learning rate of 0.001 and a batch size of 4. Both the training and testing are carried out on a desktop with an Nvidia GeForce 3090 GPU. \Revised{The total size of our network is around 21.82$M$.} As for time efficiency, in the testing stage, when we set $D=1$, the forward pass takes around 12$ms$.

%%%%%%%visual general shape%%%%%%%%%%
\begin{figure*}[t]
\centering
\begin{overpic}[width=0.97\textwidth]{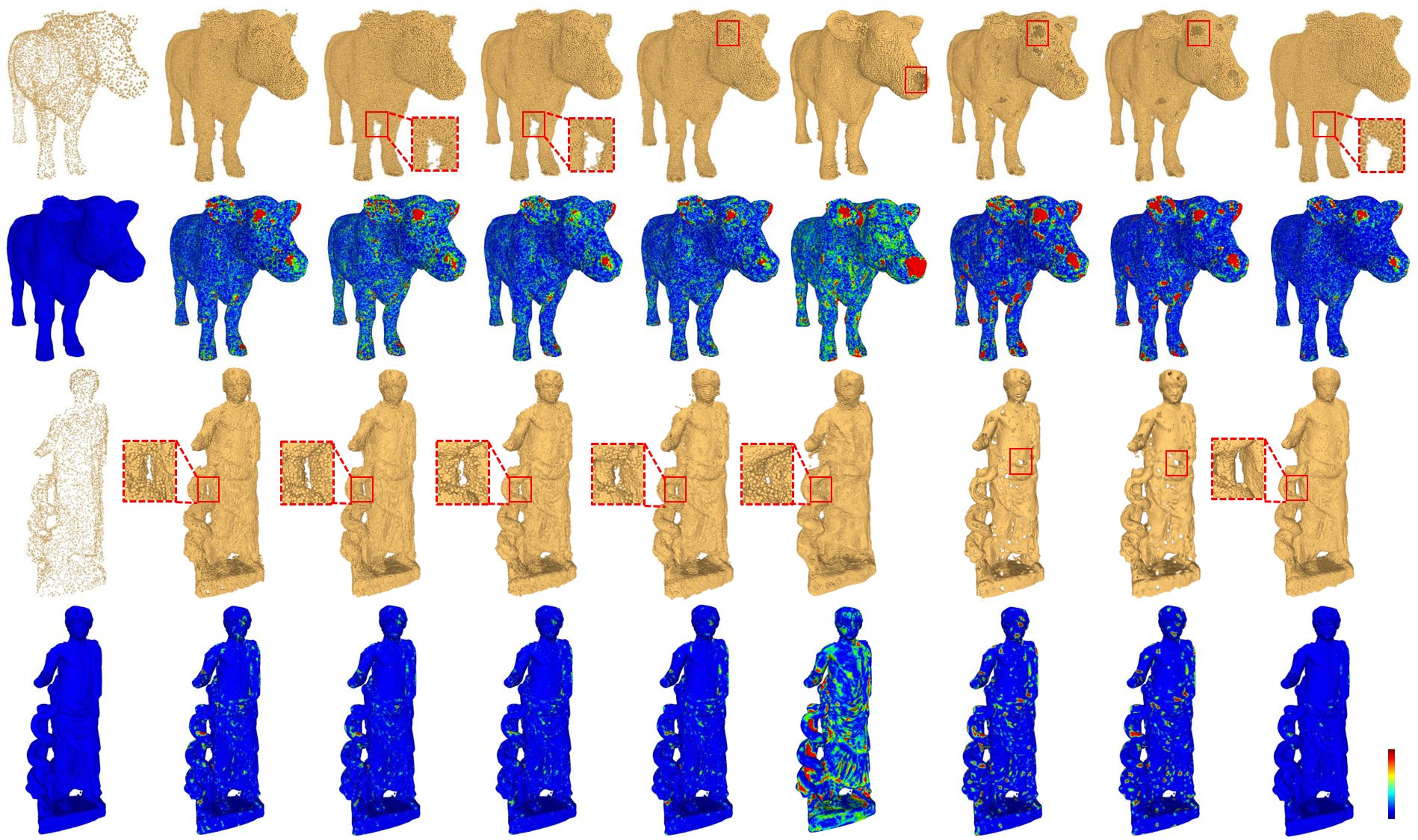} % grid,tics=2,
    \put(0, -2) {\small (a) Input \& GT}
    \put(12, -2) {\small (b) PUGAN}
    \put(23.5, -2) {\small (c) MPU}
    \put(34, -2) {\small (d) Dis-PU}
    \put(44, -2) {\small (e) PUGeo-Net}
    \put(56.5, -2) {\small (f) APU}
    \put(67, -2) {\small (g) MAFU}
    \put(76.5, -2) {\small (h) Neural Points}
    \put(89.5, -2) {\small (i) Ours}
    \put(97.2, -0.4) {\scriptsize 0.1}
    \put(97.2, 8.2) {\scriptsize 0.8}
    \end{overpic}
\caption{\textbf{Visual Comparison with $\times16$ Upsampling.} Each model has two rows, the first row shows the upsampled results, while the second row shows the color-coding of the Chamfer distance from the ground truth to the upsampled points. The biggest error appears around the holes (\textcolor{red}{red}), while the noises and outliers are usually depicted with \textcolor{green}{green}. Our results are closer to \Revised{blue}, as in the ground truth.}
\label{fig:comp_general}
\end{figure*}
%%%%%%%%%%%%%%%%%%%%%%%%%%%%%%%%%%

%%%%%%%visual CAD %%%%%%%%%%%%
\begin{figure*}[!t]
\centering
\begin{overpic}[width=0.97\textwidth]{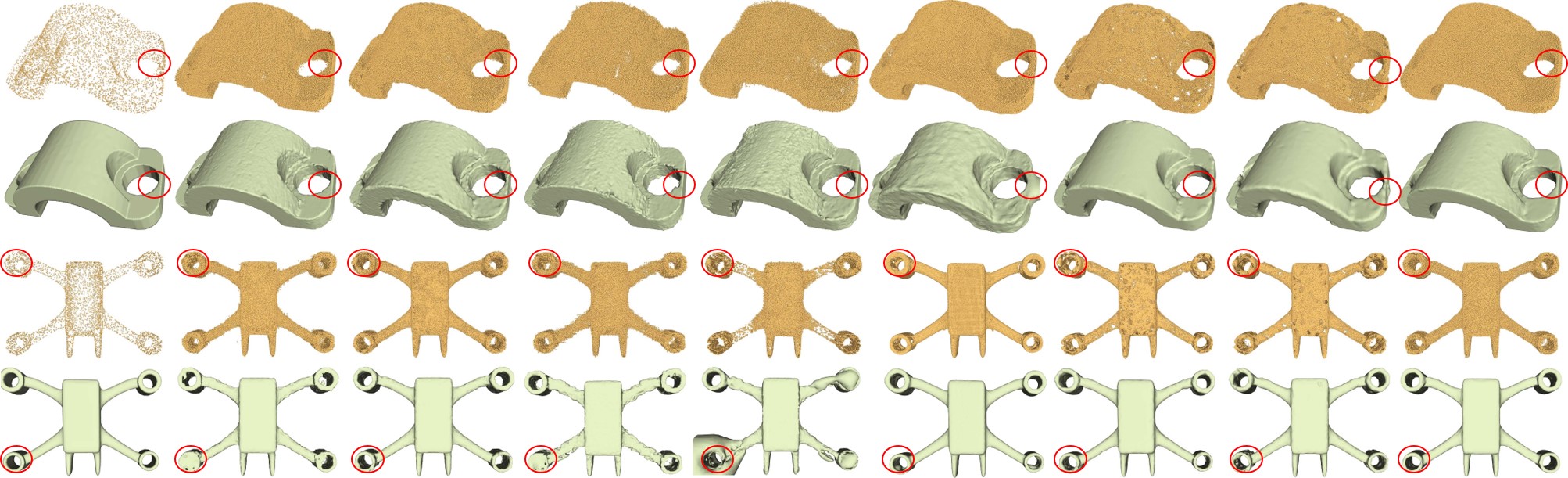} % grid,tics=2,
    \put(0, -2) {\small (a) Input \& GT}
    \put(12, -2) {\small (b) PUGAN}
    \put(25, -2) {\small (c) MPU}
    \put(34, -2) {\small (d) Dis-PU}
    \put(44, -2) {\small (e) PUGeo-Net}
    \put(58.5, -2) {\small (f) APU}
    \put(68, -2) {\small (g) MAFU}
    \put(78, -2) {\small (h) Neural Points}
    \put(92, -2) {\small (i) Ours}
    \end{overpic}
\caption{\textbf{CAD Shape Upsampling ($\times16$) and Reconstruction.} For each example, we present the upsampled result in the first row, followed by the reconstructed surface in the second row.}
\label{fig:comp_cad}
\end{figure*}
%%%%%%%%%%%%%%%%%%%%%%%%%%%%%%%%%%

%%%%%%%statistical comparison%%%%%
\begin{table*}
\caption{\textbf{Statistical Comparison.}
\Revised{Four upsampling ratios are used, and we report CD($\times10^{-4}$), HD($\times10^{-3}$), P2F($\times10^{-3}$), Uni($\times10^{-3}$) metrics, respectively. Note here, the ground truth point clouds are $P_{10k}, P_{20k}, P_{40k}$, and $P_{80k}$.}}
\label{tab:stas_comp}
\vspace{-4mm}
\centering
\subtable[Comparison on General Shape]{
\begin{tabular}{l | c c c c | c c c c| c c c c | c c c c}
		\toprule
		\multirow{2}*{Method}&\multicolumn{4}{c|}{$\times2$}&\multicolumn{4}{c|}{$\times4$}&\multicolumn{4}{c|}{$\times8$}&\multicolumn{4}{c}{$\times16$}  \\ 
		            \cmidrule(lr){2-17}
		               & {CD$\downarrow$}  & {HD$\downarrow$}
		               & {P2F$\downarrow$} & {Uni$\downarrow$}& {CD$\downarrow$}    
		               & {HD$\downarrow$} & {P2F$\downarrow$}  & {Uni$\downarrow$}
		               & {CD$\downarrow$}    
		               & {HD$\downarrow$} & {P2F$\downarrow$}& {Uni$\downarrow$}
		               & {CD$\downarrow$}    
		               & {HD$\downarrow$} & {P2F$\downarrow$} & {Uni$\downarrow$}\\
		
		\rowcolor{blue!10}\hline
		MPU 
		& 2.242 & 2.292 & 2.248 & 13.03 & 1.233  &   2.158  &  1.936  &  3.823 & 0.761 & 2.692 & 1.924 & 0.996 & 0.432   &  2.857 &  2.065 &0.316\\
		\hline
		PU-GAN
		& 2.314 &4.266 & 2.337 & 14.34 &  1.336   &  4.361   &   2.295 &  3.716 & 0.842 & 5.173&  2.441& 1.236 & 0.444   &  8.423  &  2.631&0.341\\
		\rowcolor{blue!10}\hline
		Dis-PU           
		&2.188 & 5.368 & 3.116 & 13.48  &  1.227   &  3.732   &  2.205  & 3.517 & 0.715 &3.764 & 2.172 & 1.037 &  0.389   &   3.982  & 2.148 &0.314\\
		\hline
		PUGeo-Net
		& 2.221 & 3.742& 2.534 & 12.48 & 1.347   &  4.655   &  2.251  & 3.478& 0.811 & 5.277 & 2.249 & 0.991  & 0.434   &  6.725  & 2.437 &0.344\\
		\rowcolor{blue!10}\hline
            APU
		& 2.315 & 4.624 & 1.967 & 13.45  &  1.276   &  4.462  &   1.922 &3.614 & 0.732 & 4.453 & 2.241  & 1.206 & 0.426   &  4.228  & 2.241 & 0.355\\
		\hline
		MAFU            
		 &  2.459 &2.654 & \textbf{1.317}  & 14.69 &  1.344  &   2.312 &  1.146   & 3.713 & 0.712 & 2.326 & 1.234 &  1.218 & 0.411    &   2.382  & 1.227&0.453\\
		 \hline
             \rowcolor{blue!10}\hline
		Neural Points
		& 2.191 & 7.713 & 1.843 & 13.21  &  1.197   &  4.861   &  \textbf{0.957} & 3.511 &0.753 &4.753 & \textbf{0.916} & 1.153  &  0.417   &  5.074   & \textbf{0.883}&0.401\\
        \hline
		 Ours
		&\textbf{2.166} & \textbf{2.264} & {1.335} & \textbf{12.41} &  \textbf{1.164}   &   \textbf{1.827}  & 1.922  & \textbf{3.432}  & \textbf{0.696}  &\textbf{2.075} & 2.102  &  \textbf{0.973} &  \textbf{0.384}   &   \textbf{2.154} & 2.057 &\textbf{0.304}\\
		\hline
    \end{tabular}
    \label{subtab:general_shape}
}
\subtable[Comparison on CAD Shape]{
\begin{tabular}{l | c c c c | c c c c| c c c c | c c c c}
		\toprule
		\multirow{2}*{Method}&\multicolumn{4}{c|}{$\times2$}&\multicolumn{4}{c|}{$\times4$}&\multicolumn{4}{c|}{$\times8$}&\multicolumn{4}{c}{$\times16$}  \\ 
		            \cmidrule(lr){2-17}
		               & {CD$\downarrow$}  & {HD$\downarrow$}
		               & {P2F$\downarrow$} & {Uni$\downarrow$}& {CD$\downarrow$}    
		               & {HD$\downarrow$} & {P2F$\downarrow$}  & {Uni$\downarrow$}
		               & {CD$\downarrow$}    
		               & {HD$\downarrow$} & {P2F$\downarrow$}& {Uni$\downarrow$}
		               & {CD$\downarrow$}    
		               & {HD$\downarrow$} & {P2F$\downarrow$} & {Uni$\downarrow$}\\
		
		\rowcolor{blue!10}\hline
		MPU 
		&2.437 & 3.412 & 2.495 & 10.09 &  1.216   &   2.674  &   1.324 &3.211 & 0.742 &2.844 & 1.366 & 0.764&  0.403   & 3.174  &  1.471 &0.214\\
		\hline
		PU-GAN
		&4.272 & 6.185 & 5.537 & 12.15 &  2.315  & 5.496    &  5.445 & 3.534& 1.343 & 7.446 & 5.431 &0.847  &  0.591   &  11.23   & 5.346 &0.215\\
		\rowcolor{blue!10}\hline
		Dis-PU            
		& 2.566 & 5.981 & 1.235 & 9.873 &  1.398  &   4.192  &  1.247 & 3.426 & 0.936 & 4.266 & 1.237 &0.748 & 0.457   &   4.143  & 1.215&0.236\\
		\hline
		PUGeo-Net
		& 2.971 &4.389 & 2.671 & 10.19 &  1.466   &  3.264  &  2.436 & 3.371& 1.063 & 3.416 & 2.248 & 0.786&   0.447  &   3.645  & 2.011 &0.249\\
            \rowcolor{blue!10}\hline
            APU
		& 2.672 & 4.624 & 1.915 & 12.33 & 1.433    &  4.714   & 1.834 &  3.369 & 0.796  &4.613 & 1.572  & 0.837 & 0.428    &  4.477   & 1.571 &0.304 \\
            \hline
		MAFU            
		&3.122 & 5.217 & 1.346 & 14.57 &  1.698  &  4.445   & 1.237   &4.697 & 1.224 & 4.418& 1.235 & 0.861& 0.461   &   4.069  & 1.243&0.413\\
            \rowcolor{blue!10}\hline
		Neural Points
		& 2.609 & 6.216 & \textbf{1.153} & 11.22& 1.432   &   5.534  &   \textbf{1.112}   &4.038 &0.781 & 5.644& \textbf{1.024}&0.823 &  0.407   &  5.372   & \textbf{0.983} &0.304\\
		\hline%\rowcolor{blue!10}
		Ours
		& \textbf{2.357} & \textbf{2.966} & 1.543 & \textbf{0.979}&  \textbf{1.116}   & \textbf{2.413}    & 1.229  & \textbf{2.879} & \textbf{0.706} & \textbf{2.637} &  1.216 & \textbf{0.729}&  \textbf{0.369}   &   \textbf{3.011} & 1.131 & \textbf{0.208}\\
		\hline
    \end{tabular}
    \label{subtab:cad_shape}
}
\end{table*}

\section{Experiments and Results}
\label{sec:results}

\subsection{Dataset}
For training, we utilize the same dataset adopted by Dis-PU~\cite{lidisup}. This dataset contains 147 different objects from the benchmark dataset, which contains simple, medium, and complex 3D models. Following the protocol in previous works, 120 models are used for training and validation according to the ratio $10:1$. For testing, to better verify the effectiveness of the proposed method, we test on multiple datasets that contain varying object types, including Dis-PU~\cite{lidisup}, PUGeo-net~\cite{qianpugeonet}, and real scanned~\cite{mikeala2019,GeigerLSU13} datasets.

We adopt the conventional configuration to prepare the patch-based synthetic data. Specifically, given a 3D mesh, we first employ the Poisson disk sampling~\cite{CorsiniCS12} on mesh faces to generate a dense and uniform point cloud (around $100k$ points, denoted as $P_{base}$) as the basis for data creation. For training data, we crop $P_{base}$ into 100 overlapped patches, and each patch occupies around $5\%$ area of the whole surface (around $10k$ points). We obtain the 100 seed points using \Revised{farthest point sampling (FPS)} on $P_{base}$ starting with a random seed index. We then use FPS to downsample each patch with 4096 points ($P_{4096}$) as the ground truth points, and further randomly sample 256 points ($P_{256}$) from $P_{4096}$ as the input point cloud, producing 12000 training pairs $\{(P_{256}, N_{256}, P_{4096})\}$, where $N_{256}$ are obtained by simply projecting back the 256 points to the 3D mesh with ground truth normals.

\vspace{1mm}
\noindent\textbf{CAD dataset.} To further verify the property of sharp feature alignment of our method, we construct a new CAD dataset with models from the ABC dataset~\cite{KochMJWABAZP19} and the Fusion360 dataset~\cite{WillisPLCDLSM21}. The new dataset contains 90 models with sharp features, 80 of them are used for training and validation ($10:1$), while 10 models are used for testing. We use a similar process as stated above to prepare the patch-based training data.

\vspace{1mm}
\noindent\textbf{Full shape upsampling and testing data.}
Although our method runs in the per-patch style, given an input sparse full shape, we can crop several overlapped patches and each has 256 points as input, and we simply merge all upsampled points from all patches. Because there is an overlap between neighboring patches, we then use the FPS to obtain the point cloud of the complete model with the desired number of points consistent with the user specification. 

Especially, since a single patch is less interesting than a full shape, we compare the upsampling quality in terms of the whole shape. As a result, we prepare the synthetic testing data as in the following. Given a testing model, similarly, we first get $P_{base}$, and downsample it with $m*r$ points (i.e., $P_{m*r}^t$, considering the input ratio $r$) as the ground truth, while further randomly downsample $P_{m*r}^t$ with $m=5000$ points (i.e., $P_{5000}^t$) as the input throughout this paper.

\vspace{1mm}
\noindent\textbf{Evaluation metrics.}
To quantitatively evaluate the performance, we consider three commonly-used evaluation metrics: the Chamfer distance (CD), the Hausdorff distance (HD), and the point-to-surface distance (P2F), the lower the better. The three metrics care about different aspects of the quality, i.e., CD measures the average quality, HD reflects the holes, noises, and outliers, while the P2F only finds the closest point on the surface for each resulting point. Even though there are holes, P2F can still be very small.

\Revised{
Inspired by the CVT energy \cite{du1999centroidal} used in geometry processing literatures, we have designed a metric (dubbed Uni) to evaluate the uniformity of point clouds. Specifically, given the upsampled point cloud $Y=\{ p_j \}_{j=1}^{M}$, we first prepare another far more dense point cloud $Y_{d}$ by Poisson disk sampling from the ground truth mesh. For each point in $Y_d$, we project it to $Y$ to find the closest point such that each $p_j$ will obtain a set of points $P_j$ that are closest points in $Y_d$. We calculate the Uni metric as:
\begin{equation}
    Uni=\frac{\sum_{j=0}^{M}\sum_{p_k \in \textbf{P}_{j}} \| p_k - p_j \|_2}{M}
\end{equation}

Here, $Y_d$ contains a fixed $800000$ point for each shape, and no matter what is the sampling ratio, we use the same $Y_d$ as the ground truth reference. Intuitively, $Y_d$ is dense enough and serves as a close approximation of the ground truth shape. The metric will obtain the lowest value if and only if $Y$ distributes uniformly around the surface.
}

\subsection{Comparisons}
To demonstrate the effectiveness of our method, we compare it with state-of-the-art upsampling methods, including PU-GAN~\cite{lipugan}, MPU~\cite{wangpatch}, MAFU~\cite{9555219}, Dis-PU~\cite{lidisup}, PUGeo-Net~\cite{qianpugeonet}, Neural Points~\cite{fengneural} and APU~\cite{dell2022arbitrary}. We retrain their networks from scratch on our datasets with their default configurations. For testing, the input shape has 5000 points, and four sampling ratios (i.e., 2, 4, 8, and 16) are used. The visual and statistical results are presented in the following.

\vspace{1mm}
\noindent\textbf{Results on general shapes.}
The first comparison is based on the dataset adapted from Dis-PU with general shapes.
Fig.~\ref{fig:comp_general} displays the visual comparison of $\times16$ upsampling, where for each model the top row shows the upsampled shape, while the second row visualizes the color-coding of Chamfer distance from the ground truth to the upsampled points, where red indicates a large error and blue indicates a small error.
As can be seen, PU-GAN suffers from severe noises and outliers (e.g., the area with \textcolor{green}{green} color-coding), as well as holes (i.e., the area with \textcolor{red}{red} color), while MPU performs better than PU-GAN but it still has some noises (see the red boxes). 
Dis-PU obtains better results than PU-GAN and MPU, but it tends to produce noises between closing parts (see the red boxes), and it cannot separate close parts clearly.
In addition, although PUGeo-Net, MAFU, and Neural Points can generate more compact and less noisy results, they are all sensitive to the uniformity and sparsity of the input, thus resulting in some small holes (see the \textcolor{red}{red} areas). Note here, regardless of the holes, Neural Points generates the most clean and compact results, which is also reflected in the lowest P2F metric in Table~\ref{tab:stas_comp}. \Revised{The results of APU are relatively uniform, but there are still some outliers (see the red boxes).}
As a comparison, our method clearly outperforms all the others, and thanks to the iterative updating strategy, our results are more uniform without obvious holes and more compact with fewer artifacts (e.g., noises and outliers), see the upsampled points and the color-coding for an illustration.

Table~\ref{subtab:general_shape} shows the quantitative comparison against state-of-the-art methods, where four upsampling ratios are used and \Revised{four} metrics are reported.
Overall, considering the CD, HD, \Revised{Uni} metrics, our method outperforms all the others, which is consistent with the qualitative examples.
As for the P2F metric, MAFU and Neural Points improve significantly over MPU, PU-GAN, Dis-PU, and APU, \Revised{while ours are higher than MAFU and Neural Points}, and Neural Points gets the lowest error.
\Revised{The Uni metric measures the uniformity of the upsampled point cloud. Thanks to the iterative updating scheme, our method achieves the best results.}
Another observation here is that a higher sampling ratio results in upsampled point cloud with higher quality, e.g., the CD metrics of our method for the increasing ratios are 2.166 vs. 1.164 vs. 0.696 vs. 0.384, respectively. This makes sense because the geometry of the desired shape can be expressed better by more and more points, and the same finding appears on the CAD dataset.

%%%%%%%visual Recons. %%%%%%%%%%%%
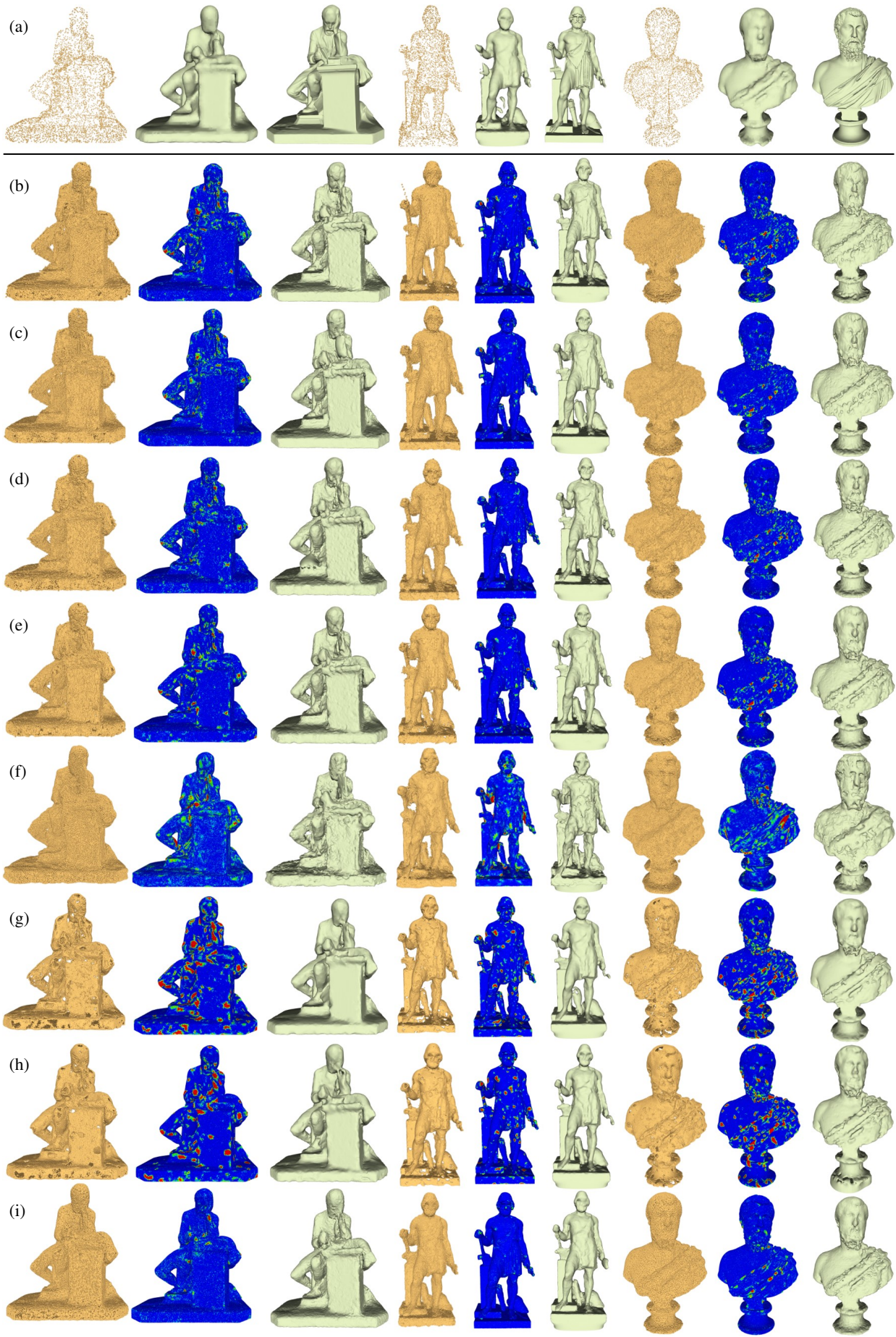
\begin{figure*}
\centering
\begin{overpic}[width=0.9\textwidth]{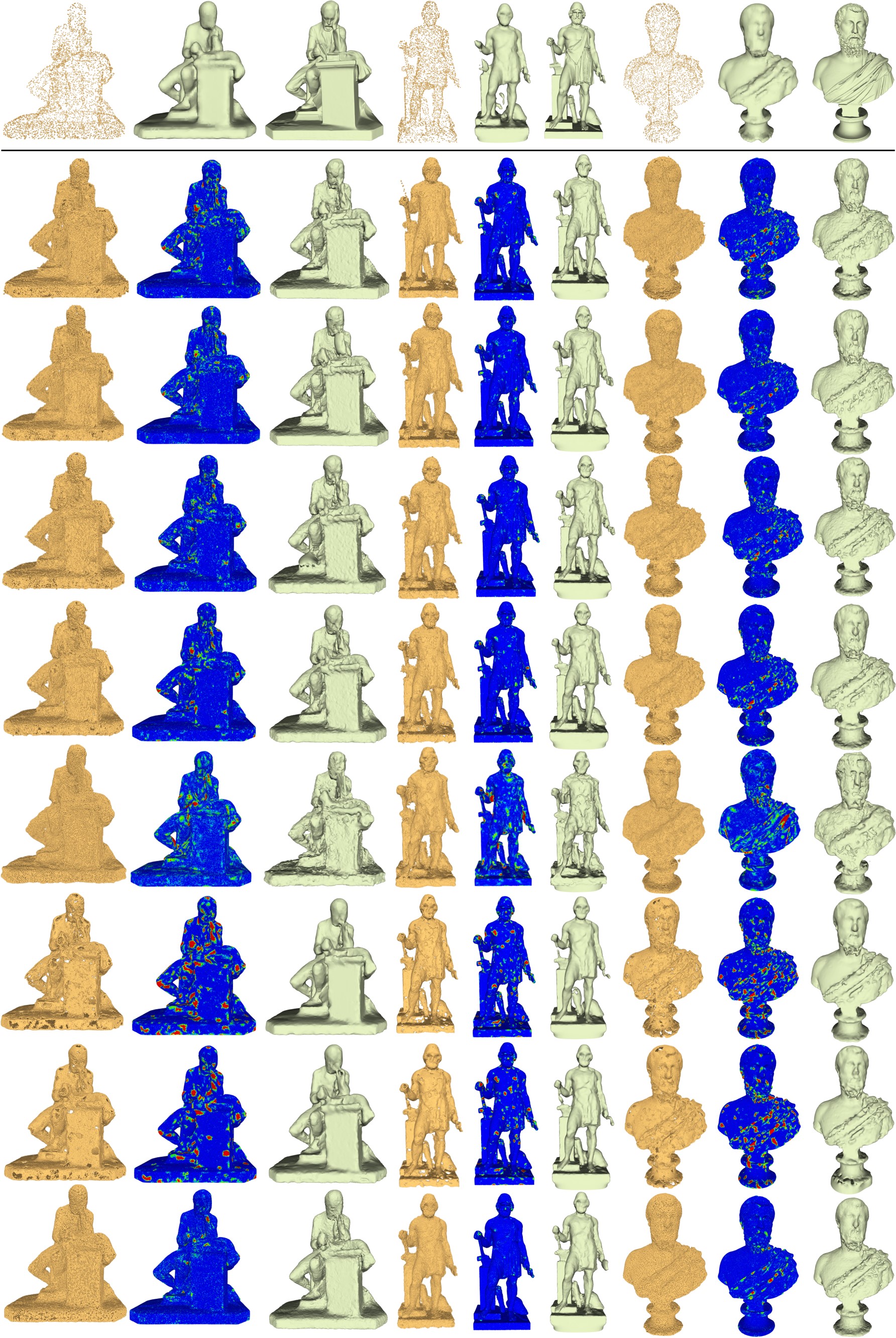} %grid,tics=2,
    \put(0.5, 98) {\small (a) }
    \put(0.5, 86) {\small (b)}
    \put(0.5, 75) {\small (c)}
    \put(0.5, 64) {\small (d) }
    \put(0.5, 53) {\small (e)}
    \put(0.5, 42) {\small (f)}
    \put(0.5, 31) {\small (g) }
    \put(0.5, 20) {\small (h)}
    \put(0.5, 9) {\small (i)}
    %\put(2, -1) {\small Result}
    \end{overpic}
    \caption{\textbf{Surface Reconstruction with $\times16$ Upsampling.} \Revised{Standard Possion reconstruction algorithm is used and if there is no predicted normal, we simply apply PCA normal estimation. (a) Input, the reconstruction from input, and GT, (b) PUGAN, (c) MPU, (d) Dis-PU, (e) PUGeo-Net, (f) APU, (g) MAFU, (h) Neural Points, (i) Ours.}}
    \label{fig:comp_recons}
\end{figure*}
%%%%%%%%%%%%%%%%%%%%%%%%%%%%%%%%%%

\Revised{Besides, we further evaluate the upsampling quality using the reconstructed surface on complex shapes. As can be seen in Fig. \ref{fig:comp_recons}, in the first row, we show the input point cloud, the direct reconstructed surface from the input, and the corresponding ground truth surface for the three shapes. In the following rows, we display the results from all competitors and ours. For each example, we show the upsampled point cloud, the Chamfer distance color-coding as explained in Fig. \ref{fig:comp_general}, and the reconstructed surface. For the reconstruction, we employ the standard Poisson reconstruction algorithm \cite{KazhdanH13}. If the method does not predict normals, we simply apply PCA normal estimation on the point cloud.

Visually, all the competitors can be roughly divided into two groups, i.e., $G_1$=$\{$MPU, PU-GAN, Dis-PU, PUGeo-Net, APU$\}$ and $G_2$=$\{$MAFU, Neural Points$\}$. The methods in $G_1$ tend to produce results with larger noises and outliers, which can be seen from the large portion of green colors. The noises and outliers can also be seen from the bumpy surfaces after the Poisson reconstruction.
The methods in $G_2$ produce high-fidelity points with fewer portions of green colors, and the reconstructed shapes have clean and smooth surfaces. However, they failed to fill the gaps due to the nonuniformity and sparsity of the input (see the red colors in the color-coding). It is worth mentioning that Poisson reconstruction is robust to small holes, thus it is not visible from the reconstructed surface solely. Our results overcome these drawbacks simultaneously, i.e., compared with methods in $G_1$, we are less noisy and more uniform, while compared with methods in $G_2$, we can fill in the holes successfully. In terms of noise, our method is in between $G_1$ and $G_2$ but is closer to $G_2$, and the P2F metric reflects this comparison. A separate P2F loss or a smoothness loss may help us, which we leave for future work.

Compared with the ground truth surfaces, all methods lose fine-grained geometric details. The reasons for us come from two aspects. The first aspect is the sparsity of the input point, which means there are no geometric details in the input (see the input point cloud and the reconstructed shape in Fig. \ref{fig:comp_recons} (a)). As long as more input points are provided, we can successfully reconstruct more details. The second aspect is the iterative updating strategy, which smooths geometric features to some extent. Although we encourage a larger outer iteration number at testing time, there should be a trade-off between detail preservation versus denoising (see the ablation study later for this trade-off). Losing geometric features can be seen as a drawback of our iterative updating strategy, but we argue that this strategy is more practically useful since the captured points are often noisy, and anti-noise robustness is more important for such point cloud processing algorithms.
}

\vspace{1mm}
\noindent\textbf{Results on CAD shapes.} Guided by the cross field, our method is sharp feature alignment. We further validate it via the second comparison on CAD shapes. Qualitative and quantitative results are shown in Fig.~\ref{fig:comp_cad} and Tab.~\ref{subtab:cad_shape}.

Qualitatively, we demonstrate two typical examples in Fig.~\ref{fig:comp_cad}, where each model has two rows: the upsampled point cloud and the reconstructed shape.
Similar observations as in Fig.~\ref{fig:comp_general} are found, i.e., MAFU and Neural Points produce less noisy results but with some holes, while the results of the remaining \Revised{four} methods often have noises and outliers, which leads to inaccurate normals used in the surface reconstruction.
It is worth noting that around the sharp feature areas, all the competitive methods do not perform well (see the highlighted red boxes), which results in blurred sharp features in the reconstructed surface. 
Also, PUGeo-Net obtains the worst reconstruction results either with a large amount of noise or irregular parts, due to the inaccurate normal estimation from the resulting point cloud.
Instead, our method produces superior results with clear sharp features that are closer to the ground truth. 
Quantitatively, a similar conclusion can be derived as shown in Table~\ref{subtab:cad_shape}. Our results outperform most of the competitors in terms of CD, HD, and \Revised{Uni}, while we have a \Revised{bigger} error on P2F compared with Neural Points.

%%%%%%%%%%%%%%%%%Self-Sampling%%%%%%%%%%%%%%%%
\begin{figure}[!t]
    \centering
    \begin{overpic}[width=0.97\linewidth]{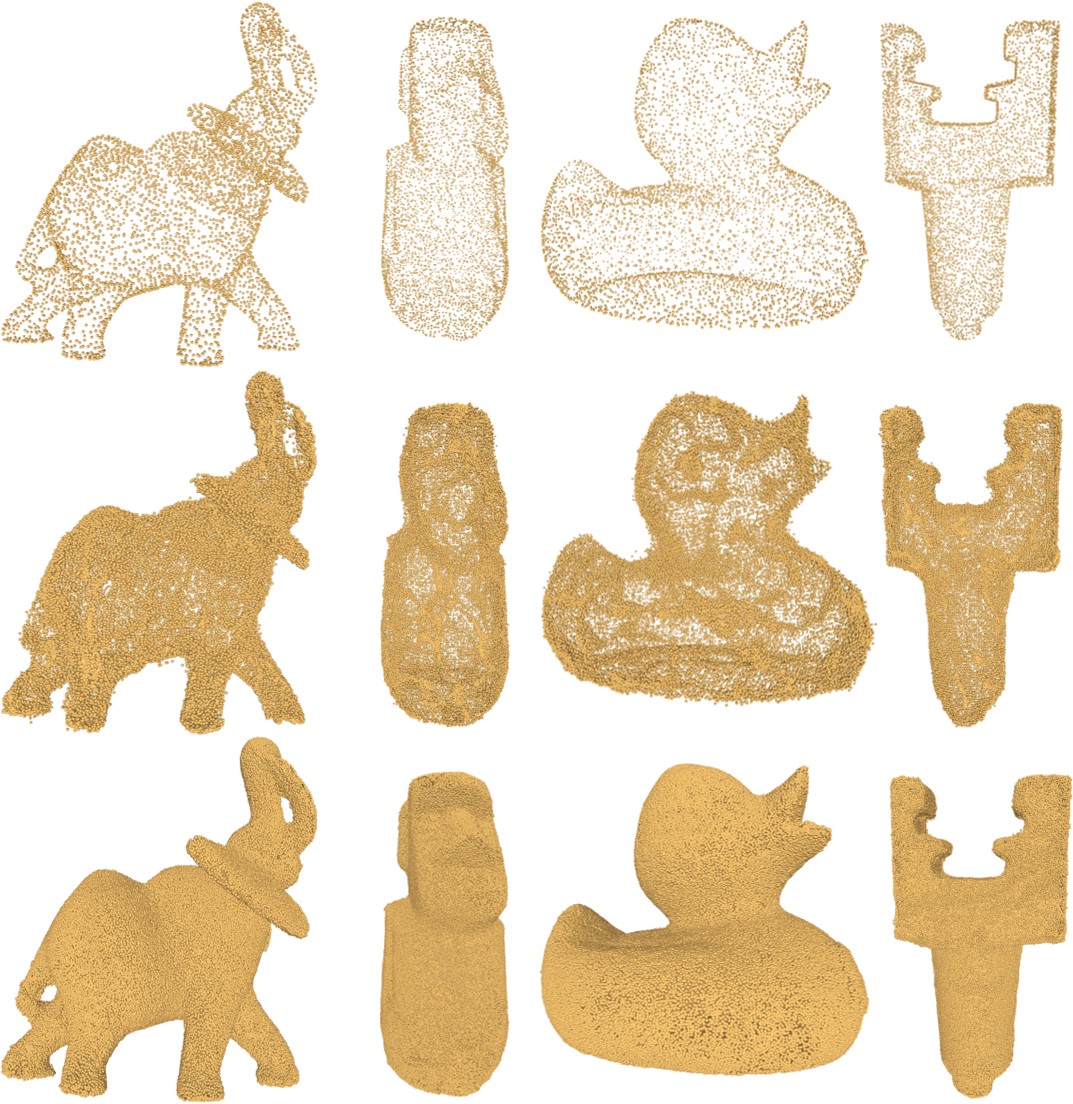} % grid,tics=5,
    \put(2, 95) {\small (a) Input}
    \put(2, 60) {\small (b) \cite{metzer2021self}}
    \put(2, 27) {\small (c) Ours}
    \end{overpic}
    \caption{\textbf{Visual Comparison with \cite{metzer2021self}}. The input contains 5000 points, and we upsample it with $r=16$.}
    \label{fig:comp_self}
\end{figure}
%%%%%%%%%%%%%%%%%%%%%%%%%%%%%%%%%%%%%%%%%%%%%%%%

\vspace{1mm}
\noindent\textbf{Comparison with \cite{metzer2021self}}. 
Metzer \etal~proposed an unsupervised point cloud consolidation method, where they also iteratively update the positions of a subset of the input points according to different criteria, e.g., denoising or upsampling. We thus compare our method with \cite{metzer2021self} on the upsampling task. To conduct a fair comparison, given an input point cloud with 5000 points, we run their experimental code with {\emph all} their criteria and upsample the input into $\times4$ and $\times16$ point clouds, respectively.
As can be seen in Fig.~\ref{fig:comp_self}, their method produces results with a large amount of noise, this may be caused by the sparsity of the input. Instead, our results are clean and compact, and clearly superior to theirs.

%%%%%%%%%%%%%%%% Real Scan %%%%%%%%%%%%%
\begin{figure*}[!t]
\centering
\begin{overpic}[width=\linewidth]{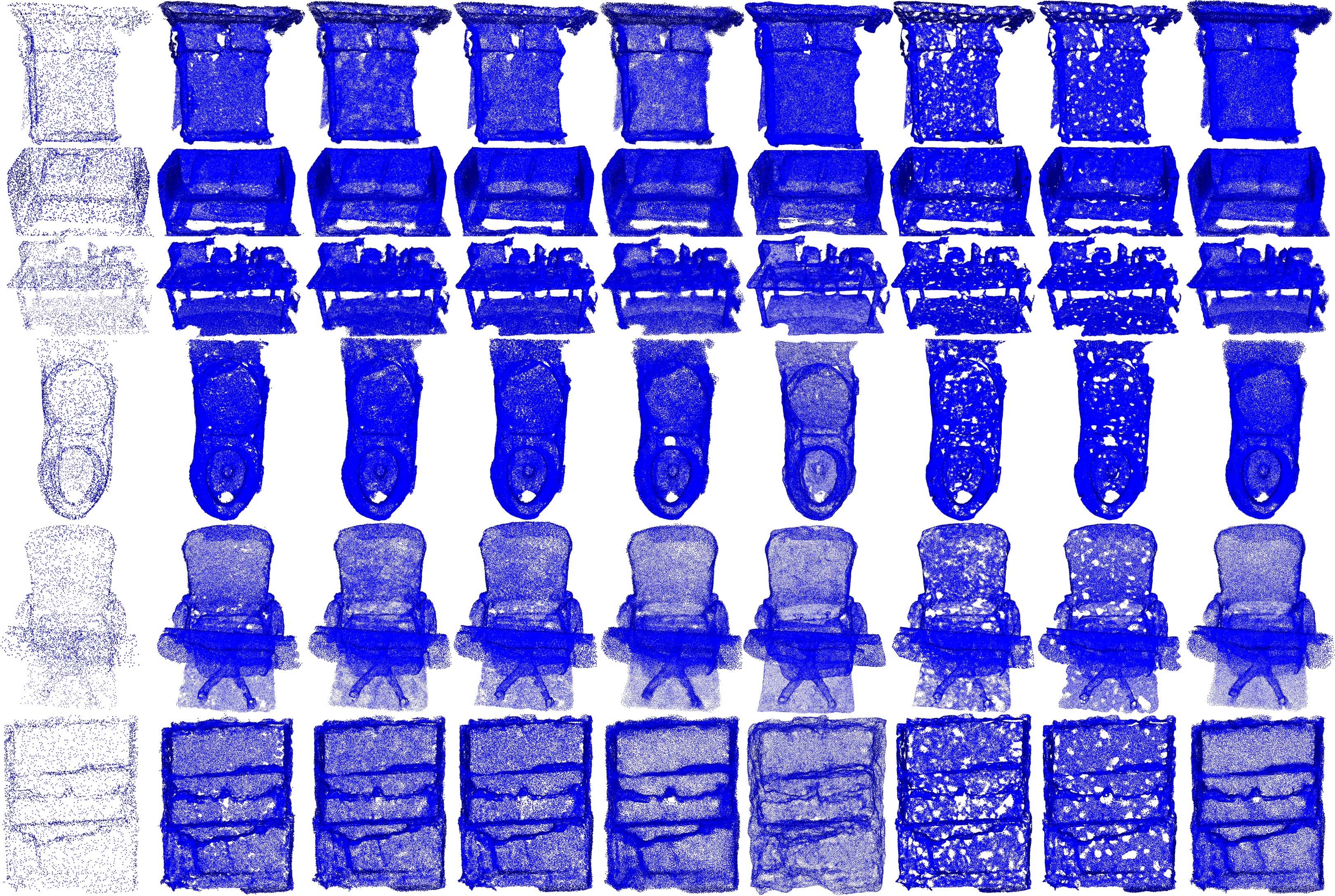} % grid,tics=2,
   \put(0, -2) {\small (a) Input}
    \put(13, -2) {\small (b) PUGAN}
    \put(25, -2) {\small (c) MPU}
    \put(34, -2) {\small (d) Dis-PU}
    \put(44, -2) {\small (e) PUGeo-Net}
    \put(58, -2) {\small (f) APU}
    \put(68.5, -2) {\small (g) MAFU}
    \put(78, -2) {\small (h) Neural Points}
    \put(92, -2) {\small (i) Ours}
    \end{overpic}
\caption{\textbf{Upsampling ($\times16$) on Real-world Scans}. \Revised{We compare our method with state-of-the-art methods on real-world scans proposed by \cite{mikeala2019}. Although our results are not perfect, rich geometric details can be seen and they are significantly superior to others. }}
\label{fig:real_general}
\end{figure*}
%%%%%%%%%%%%%%%%%%%%%%%%%%%%%%%%%%%%%%%%%%%%%%%%%%%%

%%%%%%%%%%%%%%%% LiDAR Scan %%%%%%%%%%%%%%%%%%%%%%%%
\begin{figure}[!t]
    \centering
    \begin{overpic}[width=\linewidth]{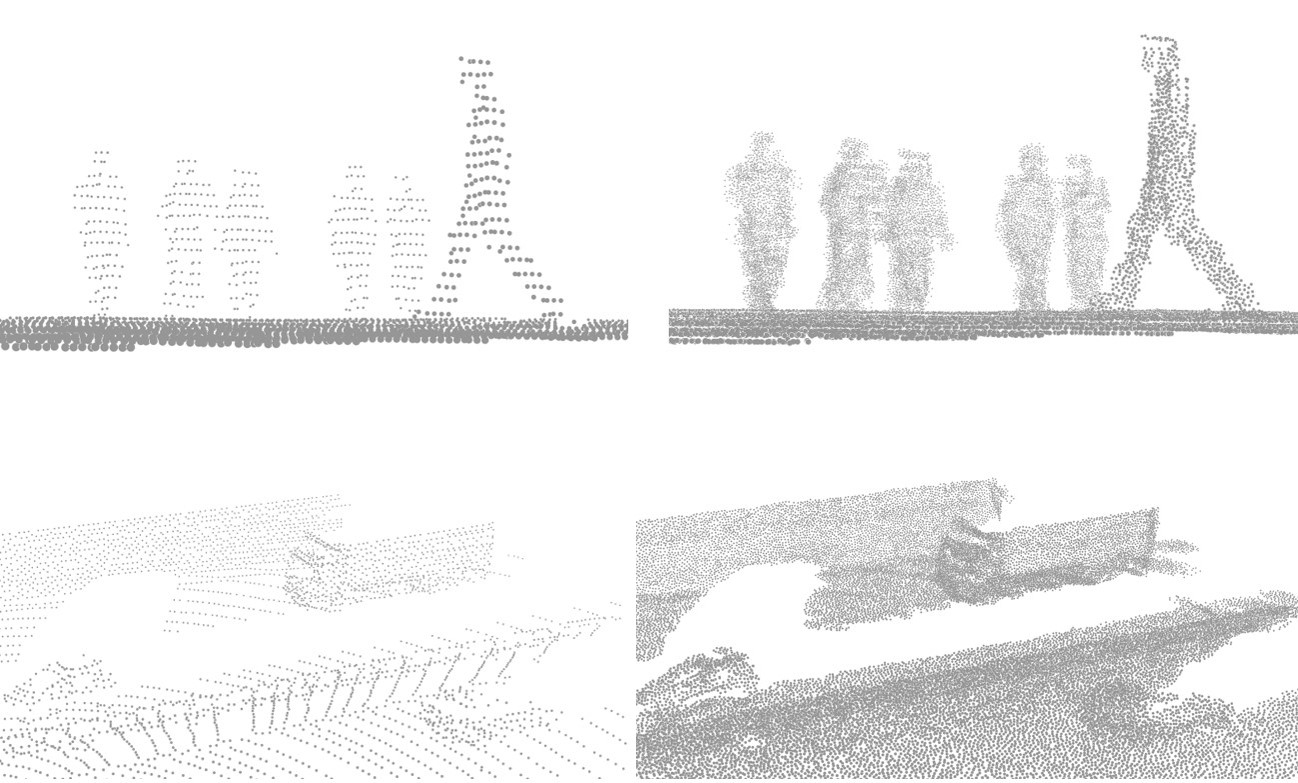}
        \put(18,-3.5) {\small (a) Input}
        \put(58,-3.5) {\small (a) Upsampled Points}
    \end{overpic}
    \caption{\textbf{Upsampling ($\times16$) on LiDAR Scan.} Even if the input contains a stripe pattern due to the scanning devices, our method can upsample the point cloud to be uniform and clean.}
    \label{fig:real_kitti}
\end{figure}
%%%%%%%%%%%%%%%%%%%%%%%%%%%%%%%%%%%%%%%%%%%%%%%%%%%

\subsection{Results on Real Scans}
\label{subsec:real_data}
Although our network is trained on synthetic datasets, it can be applied to real-world scans. Thus, we have conducted two additional experiments to demonstrate the generalization ability and robustness of our method. 

Specifically, we first compare our approach with the state-of-the-art methods on the real-world scan dataset proposed by Uy \etal~\cite{mikeala2019}. We use the trained networks on our general dataset (i.e., the same networks as in the comparison in Fig.~\ref{fig:comp_general} and Tab.~\ref{subtab:general_shape}), and directly test them without any fine-tuning. Since there is no ground truth in \cite{mikeala2019}, we only present the visual comparison in Fig.~\ref{fig:real_general}. As can be seen, the input point clouds are usually incomplete, contain several holes, and are noisy as well. Methods, like PUGeo-Net, MAFU, and Neural Points, are sensitive to the input quality, leading to several holes in the results, while results from PUGAN, MPU, Dis-PU, and APU are often noisy and non-uniform. As a comparison, our results are significantly better with enough geometric details.

Besides, we further test our trained network on a more challenging KITTI \cite{GeigerLSU13} dataset captured by LiDAR devices, a visual example is shown in Fig. \ref{fig:real_kitti}. The challenge comes from the sparsity and non-uniformity (e.g., the special stripe pattern due to the capturing functionality). Even though our network does not see such kind of data, the upsampled point clouds are uniform and clean, e.g., the human and the truck are successfully consolidated. We believe the resulting dense point cloud can greatly facilitate downstream applications, e.g., autonomous driving.

\subsection{Ablation study}
\label{subsec:abl_study}

%%%%%%%%%%%%% Ablation study stas %%%%%%%%%%%%%%%%%%
\begin{table}[t]
\caption{\textbf{Statistical Results of Ablation Study}. We upsample the input point cloud with $r=16$, and test all the alternative methods on both general shape and CAD shape datasets. CD($\times10^{-4}$), HD($\times10^{-3}$), P2F($\times10^{-3}$), Uni($\times10^{-3}$) metrics are reported.}
\label{tab:abl_stas}
\centering
    \begin{tabular}{l | p{0.17in}p{0.17in}p{0.17in}p{0.17in}| p{0.17in}p{0.17in}p{0.17in}p{0.17in}}
		\toprule
		\multirow{2}*{Method}&\multicolumn{4}{c|}{General Shape}&\multicolumn{4}{c}{CAD Shape}  \\ 
		            \cmidrule(lr){2-9}
		               & CD$\downarrow$  &HD$\downarrow$
		               & P2F$\downarrow$ &Uni$\downarrow$  &CD$\downarrow$    
		               & HD$\downarrow$ & P2f$\downarrow$  &Uni$\downarrow$   \\
		\rowcolor{blue!10}\hline
		BNet
		&  0.623   &  2.921   &   3.176 & 0.462 &   0.701  &  4.011  &  4.221& 0.433\\
		\hline
		BF 
		&  0.527   &  2.537  &   2.641  & 0.346 &   0.484  &  3.553 & 3.074  & 0.251\\
		\rowcolor{blue!10}\hline
		BF+Iter$^3$
		&  0.403   &   2.428  &  2.544   &0.328 &  0.401   &  \textbf{2.934}  & 1.254 & 0.234\\
		\hline
		BF+Iter$^5$            
		&  0.393   &   2.234  &   2.145 & 0.311 &   0.376  &  3.117   & 1.147& 0.207\\
		\rowcolor{blue!10}\hline
		BF+Iter$^{10}$
		&  \textbf{0.384}   &   \textbf{2.154 } &  \textbf{2.057}   & \textbf{0.304} & \textbf{0.369}   &   3.011  & \textbf{1.131} & \textbf{0.208}\\
		\hline
    \end{tabular}
\end{table}
%%%%%%%%%%%%%%%%%%%%%%%%%%%%%%%%%%%%%%%%%%%%%%%%%%%

%%%%%%%%%%%%% w/, w/o iteration %%%%%%%%%%%%%%%%%%
\begin{figure}[t]
    \centering
    \begin{overpic}[width=\linewidth]{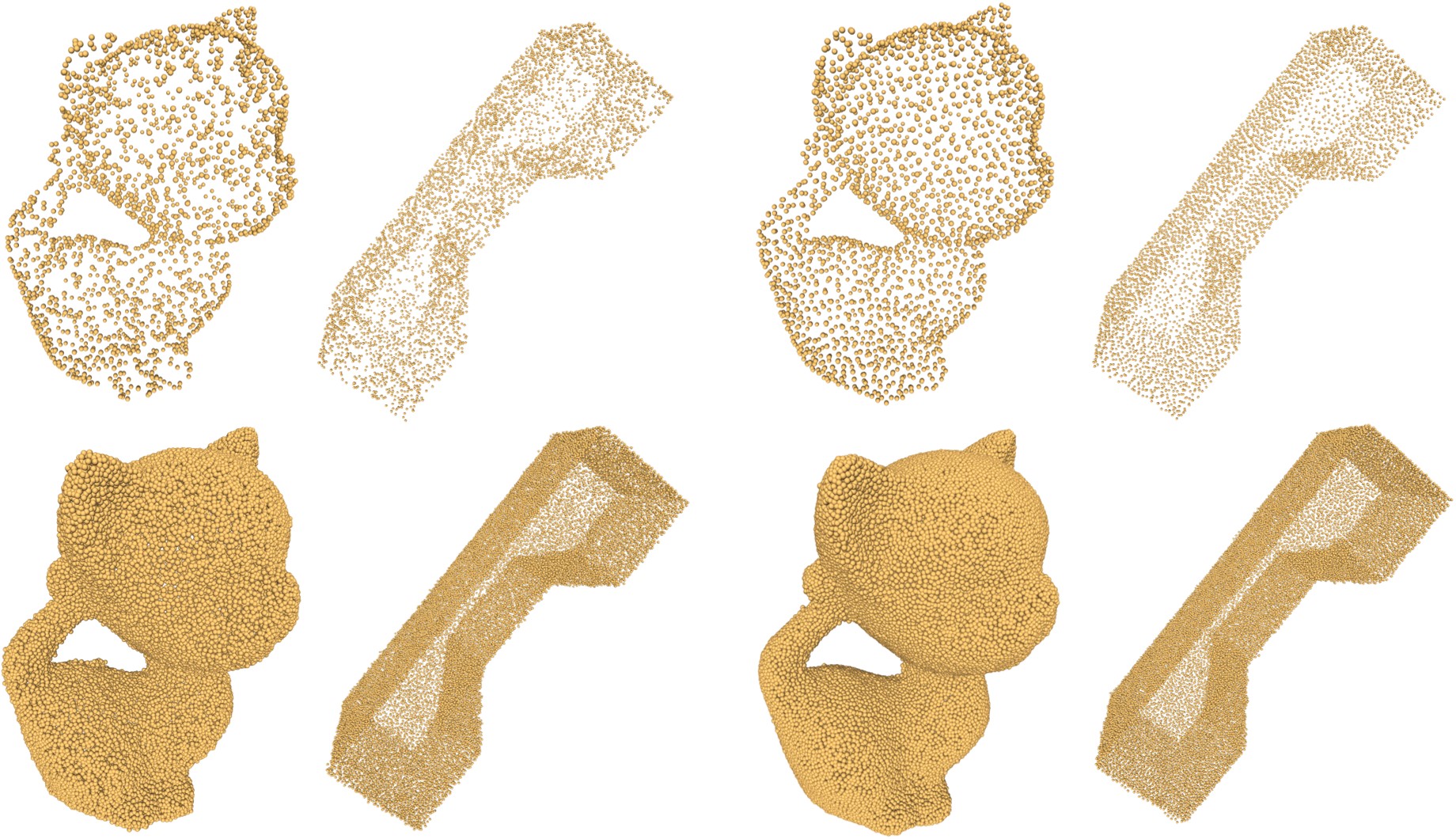} % grid,tics=5,
        \put(13, -3.5) {\small (a) BNet+Field}
        \put(66, -3.5) {\small (b) BF+Iter$^{10}$}
    \end{overpic}
    \caption{\textbf{Upsampling Results ($\times16$) with or without Iterative Updating.} The first row shows the original (i.e., BNet+Field) or updated (i.e., BF+Iter$^{10}$) input points, while the second row shows the upsampled point clouds.}  
    \label{fig:abl_vis}
\end{figure}
%%%%%%%%%%%%%%%%%%%%%%%%%%%%%%%%%%%%%%%%%%%%%%%%%%

To validate the effectiveness of the technical designs of our approach, we have conducted several ablation experiments with alternative network structures:
\begin{enumerate}
    \item BNet: our baseline network for the upsampling task, where we first predict point-wise features and normals using the GDCNN backbone, then in the mapping function learning, since there is no predicted cross-field, we randomly choose two orthogonal directions on the tangent plane, which are combined with the predicted normal to serve as the local coordinate for feature inpainting and offset prediction.
    \item BNet+Field (BF): we augment the baseline network with the feature-aligned cross field prediction, and in the mapping function learning, we use the cross field and normal as the local coordinate. Other configurations remain the same.
    \item BF+Iter$^x$: we further augment BNet+Field with our iterative updating strategy with $D$=$x$ iterations in both training and testing, here we choose $D$ with 3 (BF+Iter$^3$), 5 (BF+Iter$^5$), and 10 (BF+Iter$^{10}$), respectively.
\end{enumerate}

We train all the alternative methods from scratch on our general and CAD datasets, and at testing time, we fix the sampling ratio to 16 for quantitative and qualitative evaluations as shown in Fig.~\ref{fig:abl_vis} and Table~\ref{tab:abl_stas}.

\vspace{1mm}
\noindent\textbf{Cross field Guidance}.
Our method is built upon the learned cross field via the self-supervised learning, it is sharp feature alignment, which guides the geometric-aware local feature learning and offset prediction. From Tab.~\ref{tab:abl_stas}, compared with BNet, BNet+Field gains a dramatic decrease in the CD metric indicating a significant quality improvement. Especially on CAD shapes that contain many sharp features, the average CD metric drops $31\%$ (0.701 vs. 0.484).

\vspace{1mm}
\noindent\textbf{Iterative Updating Strategy}.
To deal with the sparsity and non-uniformity, and noises of the input and the resulting output, we propose an iterative updating strategy that drives the input points to distribute uniformly and push the noises and outliers to the target shape as closely as possible. The improved input points in return serve as the input for the next iteration, benefiting the cross-filed and normal prediction as well as the mapping function learning. In the end, both the upsampled point cloud and the input points obtain better quality. There is a hyperparameter $D$, i.e., the number of iterations, which can be set by trading off efficiency and quality.

To validate this novel scheme, we compare network variants with 3 (BF+Iter$^3$), 5 (BF+Iter$^5$), and 10 (BF+Iter$^{10}$) iterations. \Revised{Note, for a fair comparison, we stop the network training of these networks when we observe that there is no clear decrease in the validation curve, meaning that the network converged (the total iteration numbers are roughly the same).} Statistically, by applying the iterative update strategy, the quality of the results improves remarkably, i.e., 0.527 vs. 0.403, and 0.484 vs. 0.401 in terms of the CD metric. As we gradually increase the iteration number, the CD metric decreases gradually on both datasets, endorsing the effectiveness.
We also visualize two typical examples in Fig.~\ref{fig:abl_vis}, where we show both the input, the updated input, and the resulting point clouds of BF and BF+Iter$^{10}$, respectively. It can be seen clearly that after 10 iterations, both the input and the upsampled point cloud become more uniform with clear sharp edges and almost no noises.

\vspace{1mm}
\Revised{
\noindent\textbf{Fidelity vs. Denoising}.
While the iterative updating scheme pushes the upsampled points to be as uniform as possible, it may also cause the loss of geometric features for complex shapes. Thus, there is a trade-off between fidelity (detail reservation) versus denoising when choosing the number of iterations. 
We have investigated this trade-off as shown in Fig.~\ref{fig:iter_diff_noise}. In the first row, when the input point cloud contains $10k$ points but without noise, from $D=1$ to $D=5$, we achieve the best upsampling results with finer geometric details. However, if $D$ increases to $15$, some details are smoothed out. 
In the second row, where the same point cloud is corrupted with $2\%$ Gaussian noise, the upsampled point cloud has holes and noises at $D=1$, and more iterations ($D=5,15$) have to be applied so that the noises are gradually removed and the upsampled points are more clean and uniform.

%%%%%%%%%%%%%%%%%%%%%%%%diff iter (1,5,10)%%%%%%%%%%%%%%%%%%%%%%%%
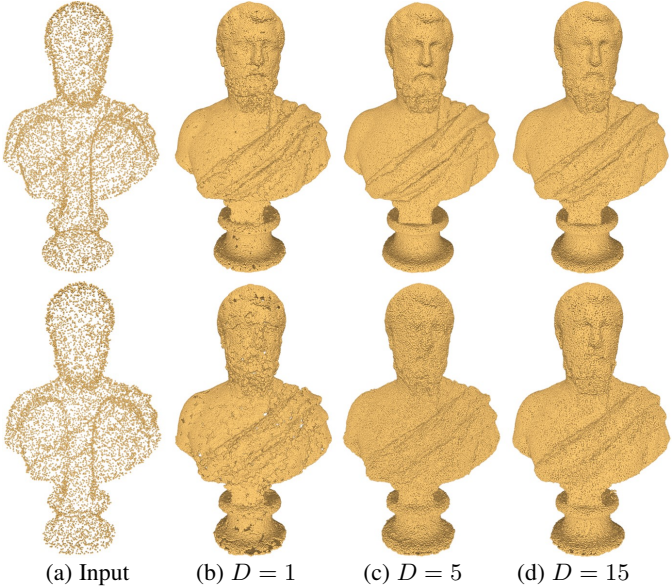
\begin{figure}[!htb]
    \centering
   \begin{overpic}[width=\linewidth]{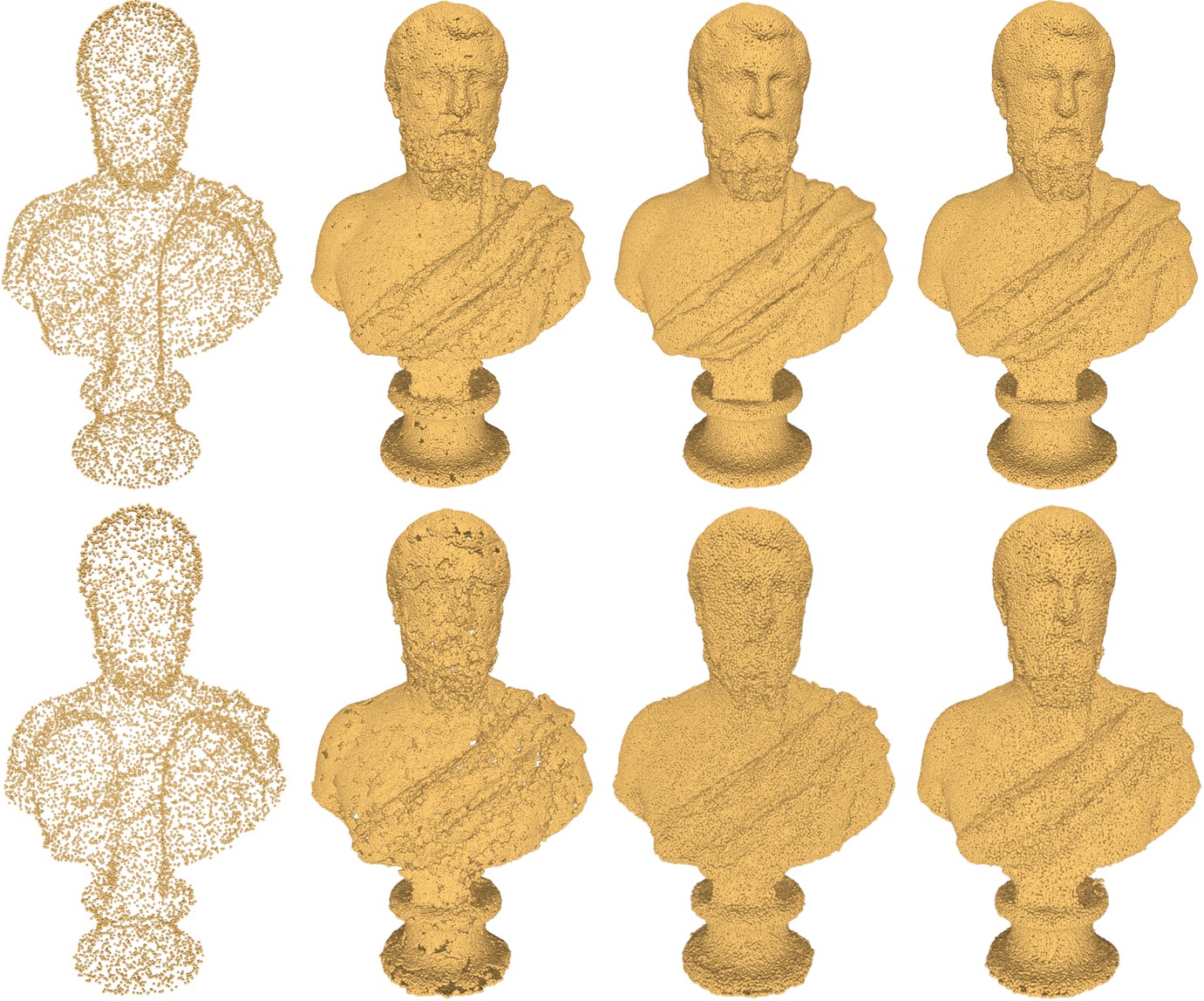}
        \put(6.5, -3.5) {\small (a) Input}
        \put(29,-3.5) {\small (b) $D = 1$}
        \put(54,-3.5) {\small (c) $D = 5$}
        \put(77,-3.5) {\small (d) $D = 15$}
    \end{overpic}
    \caption{A trade-off between fidelity (detail reservation) versus denoising.}
    \label{fig:iter_diff_noise}
\end{figure}
%%%%%%%%%%%%%%%%%%%%%%%%diff iter (1,5,10)%%%%%%%%%%%%%%%%%%%%%%%%

Generally at testing time, we suggest using a higher $D$, e.g., 10, or 15, which will bring better results because the real captured points are often noisy, and anti-noise robustness is more important in such cases.
For most of the results shown in the paper, we have used a default $D=10$ unless otherwise specified.
}

\vspace{1mm}
\Revised{\noindent\textbf{Choices of $K_1$}. $K_1$ controls the number of neighbor points when measuring the smoothness and alignment of cross-fields in Eq. \ref{Eq:fieldsmooth}. To validate its effect, we have trained three network variations of BF+Iter$^{10}$ by setting $K_1$ to 2, 6, and 16, respectively. From Tab. \ref{tab:abl_stas_k}, we can see that a smaller or bigger value results in performance degradation, because too few points lead to under-smoothed and aligned cross fields within a small area, while too many points lead to over-smoothed and even wrongly aligned cross fields by considering a larger local area.

\textbf{Choices of the voxel resolution $d$.} Fig. \ref{fig:pipeline}(b) and (c) illustrate the voxelization with the choice of $d$ clearly. Given the same spatial spacing, a smaller $d$ (e.g., 2) means a bigger voxel size and a smaller voxel number (e.g. 8), while a bigger $d$ (e.g., 10) means a smaller voxel size and a bigger voxel number (e.g., 100). When we project the selected $K_2$ neighbor points into the voxels, a smaller $d$ means a higher probability of collision of the point features, while a larger $d$ means a low occupancy rate for the point features. Both cases will affect the feature learning and further affect the offset learning of newly sampled points. Here, $d$ = 7 is determined by experimental experience.
}

%%%%%%%%%%%%% Ablation study stas %%%%%%%%%%%%%%%%%%
\begin{table}[t]
\caption{\textbf{Statistical Results of Ablation Study}. \Revised{We upsample the input point cloud with $r=16$, and test all the alternative methods on both general shape and CAD shape datasets. CD($\times10^{-4}$), HD($\times10^{-3}$), P2F($\times10^{-3}$), Uni($\times10^{-3}$) metrics are reported.}}
\label{tab:abl_stas_k}
\centering
    \begin{tabular}{l | p{0.17in}p{0.17in}p{0.17in}p{0.17in}| p{0.17in}p{0.17in}p{0.17in}p{0.17in}}
		\toprule
		\multirow{2}*{Method}&\multicolumn{4}{c|}{General Shape}&\multicolumn{4}{c}{CAD Shape}  \\ 
		            \cmidrule(lr){2-9}
		               & CD$\downarrow$  &HD$\downarrow$
		               & P2F$\downarrow$ &Uni$\downarrow$ &CD$\downarrow$    
		               & HD$\downarrow$ & P2f$\downarrow$ &Uni$\downarrow$    \\
		\rowcolor{blue!10}\hline
		$K_1$=2
		&  0.394   &  2.411   &  2.235 &0.316 & 0.378 &  3.274  &  1.336 & 0.231 \\
		\hline
		$K_1$=6
		&  \textbf{0.384}   &   \textbf{2.154 } &  \textbf{2.057}  & \textbf{0.304}& \textbf{0.369}   &   \textbf{3.011}  & \textbf{1.131} &\textbf{0.208}\\
		\rowcolor{blue!10}\hline
		$K_1$=16
		&  0.405   &   2.316  &  2.317 & 0.322& 0.391 &  3.448  & 1.405  &0.217 \\
        \hline
    \end{tabular}
\end{table}
%%%%%%%%%%%%%%%%%%%%%%%%%%%%%%%%%%%%%%%%%%%%%%%%%%%

\subsection{Robustness Analysis}
\label{subsec:analysis}
We have further conducted a series of experiments to validate the robustness of our approach by considering different aspects of our algorithm as in the following.

%%%%%%%%%%%%%%%% Field visualization %%%%%%%%%%%%%
\begin{figure}[!t]
\centering
\begin{overpic}[width=\linewidth]{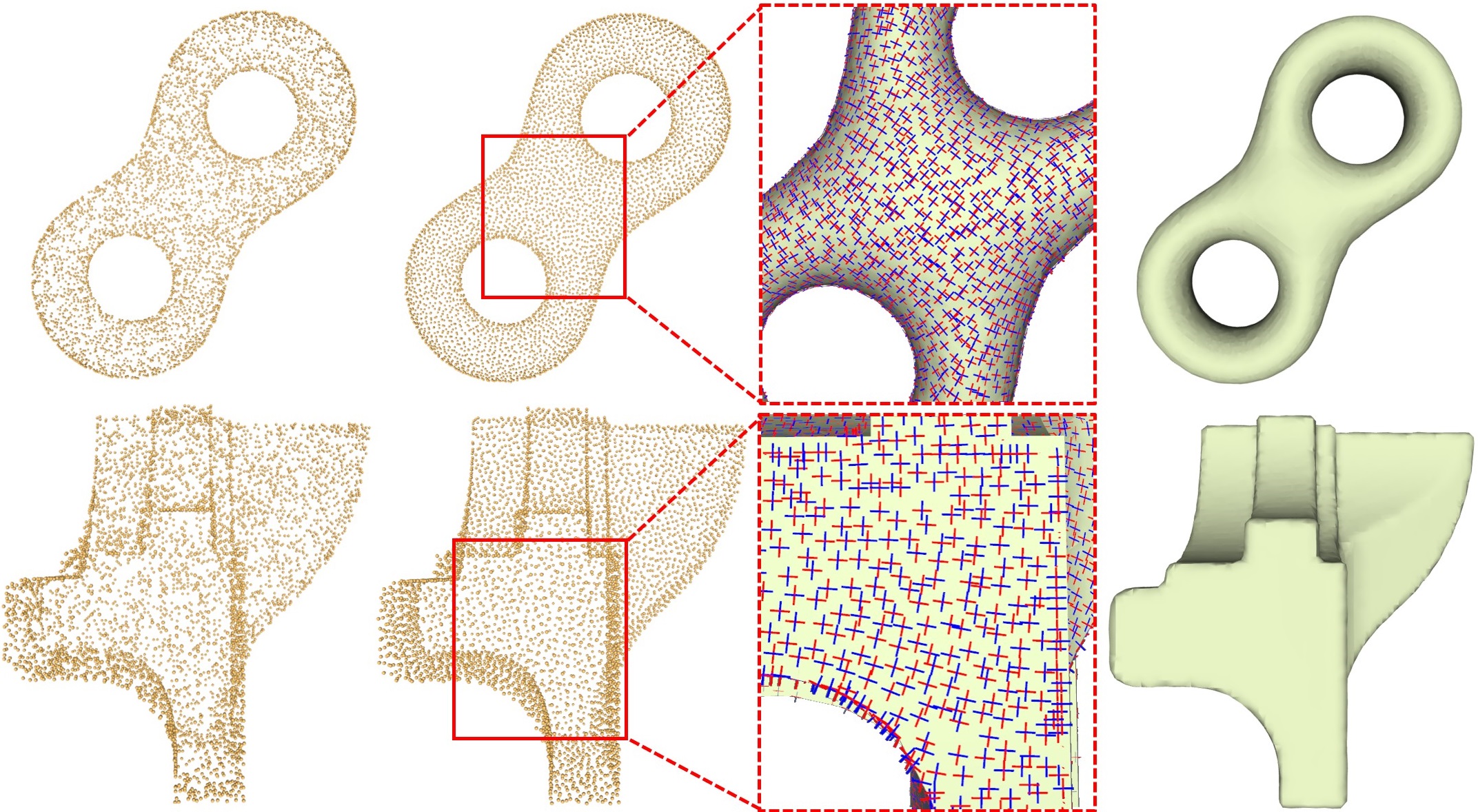}
        \put(6.5, -3.5) {\small (a) Input}
        \put(25,-3.5) {\small (b) Updated Input}
        \put(52,-3.5) {\small (c) Cross Field}
        \put(76,-3.5) {\small (d) Recons. Surf.}
    \end{overpic}
\caption{\textbf{Visual Results of Cross Field and Normal.} In the two examples, we show the (a) original and (b) updated input points, the (c) learned per-point cross field and (d) the reconstructed surface by Poisson taking as input (b) and the learned per-point normal.}
\label{fig:field_vis}
\end{figure}
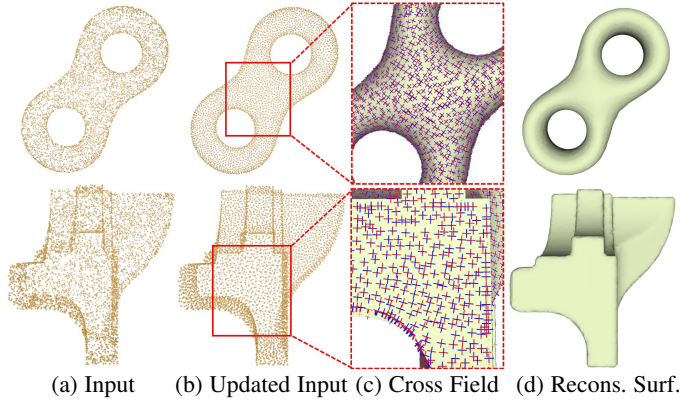
%%%%%%%%%%%%%%%%%%%%%%%%%%%%%%%%%%%%%%%%%%%%%%%%

%%%%%%%%%%%%%%%% Arbitrary Ratios %%%%%%%%%%%%%
\begin{figure}[!t]
    \centering
    \begin{overpic}[width=0.93\linewidth]{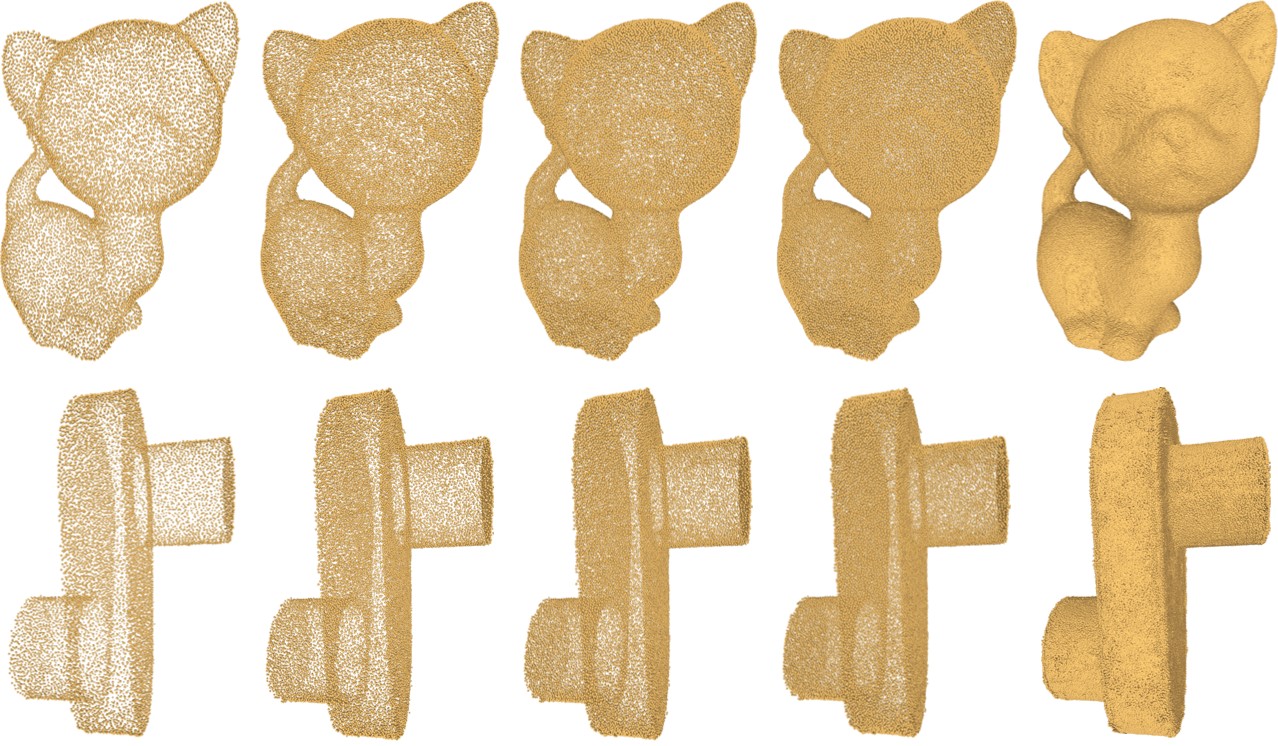}
        \put(5, -3.5) {\small $r$=1.7}
        \put(24,-3.5) {\small $r$=3.4}
        \put(44,-3.5) {\small $r$=6.3}
        \put(64,-3.5) {\small $r$=9.4}
        \put(87,-3.5) {\small $r$=60}
    \end{overpic}
    \caption{\textbf{Upsampling with Arbitrary Ratios.} Four ratios are used to upsample the same input with 5000 points, our results are all clean and uniform, and a higher ratio means a denser point cloud with more distinct geometric details.}
    \label{fig:diff_ratio}
\end{figure}
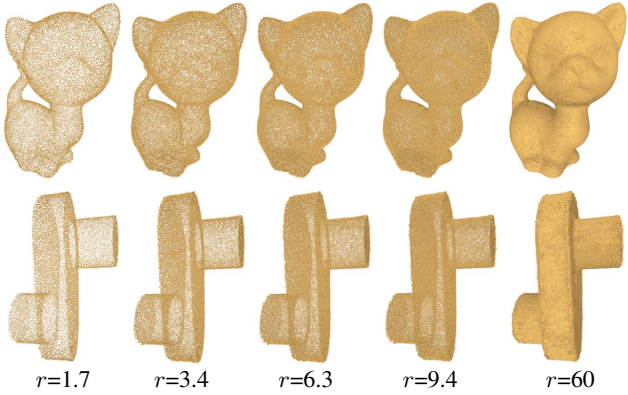%1.7 3.4 6.3 9.4
%%%%%%%%%%%%%%%%%%%%%%%%%%%%%%%%%%%%%%%%%%%%%%

%%%%%%%%%%%%%%%% Noisy Input %%%%%%%%%%%%%
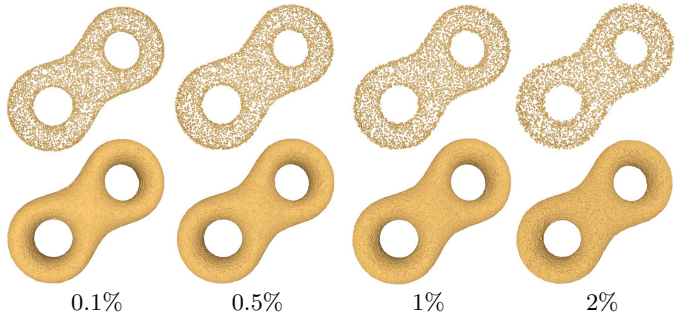
\begin{figure}[!t]
    \centering
    \begin{overpic}[width=\linewidth]{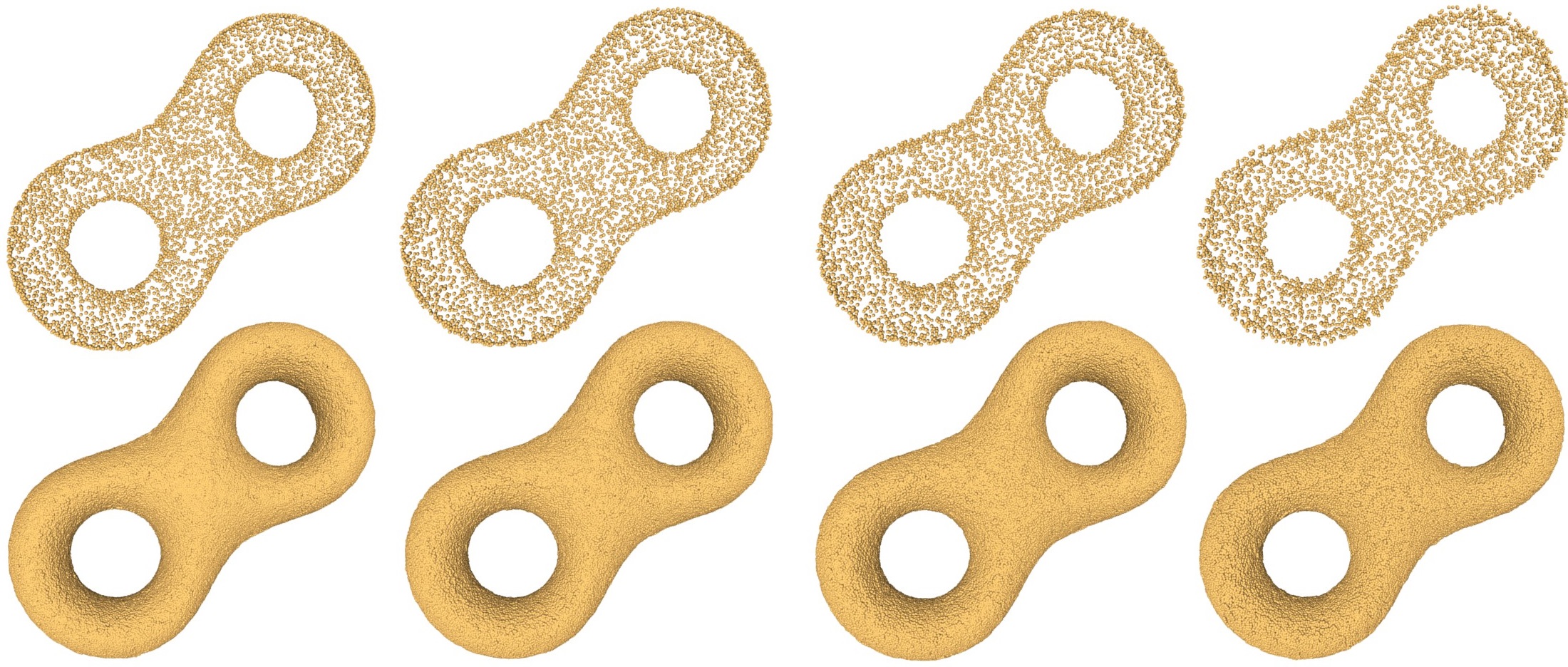}
        \put(10,-3.5) {\small $0.1\%$}
        \put(34,-3.5) {\small $0.5\%$}
        \put(61,-3.5) {\small $1\%$}
        \put(87,-3.5) {\small $2\%$}
    \end{overpic}
    \caption{\textbf{Anti-noise Upsampling ($\times16$).} The first row is input with increasing Gaussian noise levels, while the second row shows the corresponding upsampling results.}
    \label{fig:diff_noise}
\end{figure}
%%%%%%%%%%%%%%%%%%%%%%%%%%%%%%%%%%%%%%%%%%

\vspace{1mm}
\noindent\textbf{Accurate Cross Field and Normal Prediction.} 
To demonstrate the accuracy of the learned cross field and normal, we present two examples in Fig.~\ref{fig:field_vis}. The loss of cross-field learning is to let them align with sharp features and change as smoothly as possible. Also, the iterative updating strategy improves the quality of input points thus resulting in better cross-field and normal learning. We, therefore, visualize the learned cross fields by drawing them on the ground truth surface. It is clear that the fields are aligned with sharp features very well. Moreover, we reconstruct the surface from the refined and sparse input points with the learned normals, the high quality of the surface indicates the accuracy of the predicted normals.

\vspace{1mm}
\noindent\textbf{Arbitrary Upsampling.}
By the mapping function learning, our method is not restricted to specific sampling ratios. Therefore, we experiment with \Revised{five} arbitrary ratios on two examples as in Fig.~\ref{fig:diff_ratio}. Given the inputs with 5000 points, both the Kitty and the CAD mechanical part are successfully upsampled, and with a higher ratio, the results obtain more fine-grained details and clear sharp features.

\vspace{1mm}
\noindent\textbf{Anti-noise Upsampling ($\times16$).} 
Thanks to the iterative strategy, noises, and outliers are pushed back to the desired surface, thus our method is robust to input noises. We verify it by adding increasing Gaussian noises to the same input and generating the upsampled point cloud with $r=16$. As can be seen from Fig.~\ref{fig:diff_noise}, there is no clear performance degradation even with $2\%$ Gaussian noises, where the input is noise and challenging, but our results are still clean, compact, and uniform.

%%%%%%%%%%%%%%%% Input Sparsity %%%%%%%%%%%%%
\begin{figure}[!t]
    \centering
    \begin{overpic}[width=\linewidth]{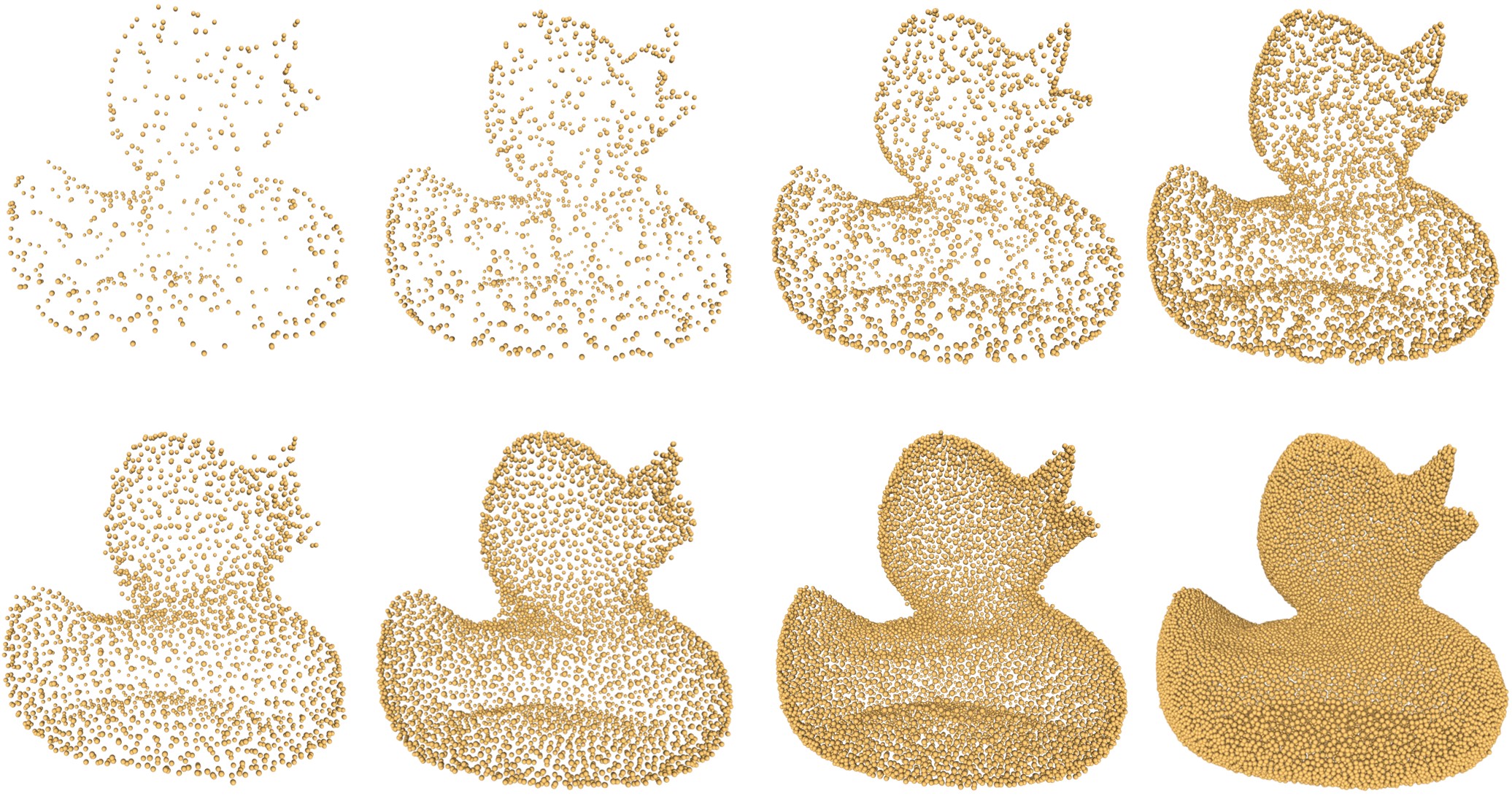}
        \put(1,49) {\small 512}
        \put(24,49) {\small 1024}
        \put(50,49) {\small 2048}
        \put(75,49) {\small 4096}
    \end{overpic}
    \vspace{-6mm}
    \caption{\textbf{Upsampling ($\times4$) with Varying Input Densities}. The first rows demonstrate different input with increasing point densities, while the second row shows the corresponding upsampling results ($\times4$).}
    \label{fig:diff_density}
\end{figure}
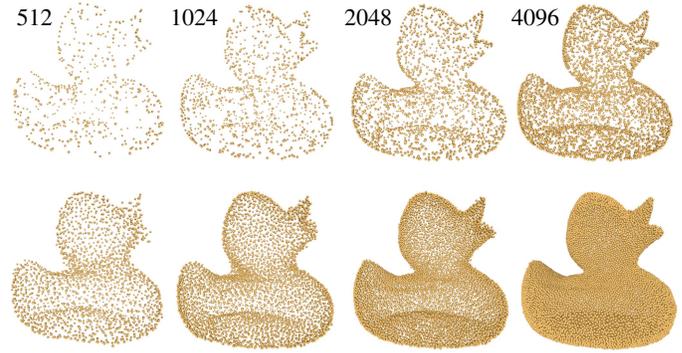
%%%%%%%%%%%%%%%%%%%%%%%%%%%%%%%%%%%%%%%%%%%

\noindent\textbf{Upsampling ($\times4$) with Varying Input Densities.}
To the best of our knowledge, existing upsampling methods are sensitive to the density of the input, because fewer points are not enough to describe the underlying geometry, thus leading to nonuniform and noisy results. However, as shown in Fig.~\ref{fig:diff_density}, we feed our method with inputs of varying densities, i.e., 512, 1024, 2048, and 4096, respectively. Even with 512 points, our method can still produce decent upsampled points depicting the desired shape with certain geometric details.

\section{Conclusion}
In this paper, we propose {\em \name}, a novel iterative algorithm for point cloud upsampling at arbitrary ratios, where we, for the first time, have introduced the point-based cross field definition and learning. The learned cross field encoding rich geometric information guides our upsampling to better capture sharp features. Besides, the newly proposed iterative updating strategy further refines the point distribution, greatly improving the result quality. Extensive evaluations demonstrate that our method is robust and outperforms state-of-the-art methods.

%%%%%%%%%%%%%%%%%%Limitations%%%%%%%%%%%%%%%%%%
\begin{figure}[!t]
\centering
    \begin{overpic}[width=\linewidth]{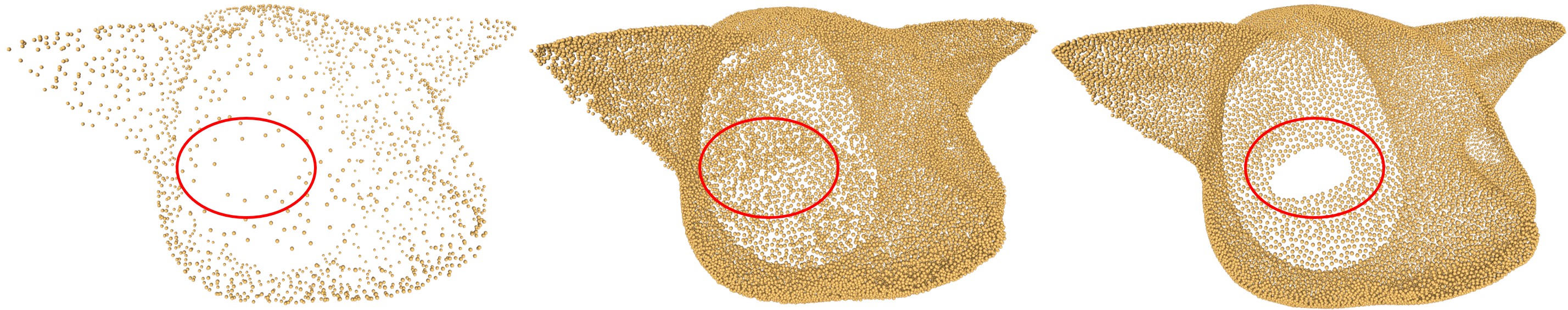}
        \put(12,-3.5) {\small (a) Input}
        \put(47,-3.5) {\small (b) Ours}
        \put(81,-3.5) {\small (c) GT}
    \end{overpic}
\caption{\textbf{Failure Case}. Given the input (a), our method wrongly fills the hole of the pig eye (b), while it should be empty in the ground truth (c).}
\label{fig:failure}
\end{figure}
%%%%%%%%%%%%%%%%%%%%%%%%%%%%%%%%%%%%%%%%%%%%%%%%%%%%%%

\vspace{1mm}
\noindent\textbf{Limitations and Future Work}.
Like most of the existing approaches, our method runs in a patch-wise manner, thus the receptive field of the network is limited, and fewer semantic and global geometric information is extracted and employed. For example, as shown in Fig.~\ref{fig:failure}, the pig eye is wrongly filled with upsampled points. From the local patch, the algorithm is hard to tell the hole is an eye, and the iterative updating scheme tends to push the input point to distribute evenly and sample more points toward this region to achieve the uniformity goal. In the future, two inspiring research directions are worth exploring, i.e., extending our method to a global setting, or further incorporating semantic information to guide the upsampling.

\ifCLASSOPTIONcaptionsoff
  \newpage
\fi

\ifCLASSOPTIONcaptionsoff
  \newpage
\fi

\section*{Acknowledgment}
The authors would like to thank the reviewers for their valuable suggestions. This work was supported by the National Natural Science Foundation of China under Grant U1909210, Grant 62172257. The work of Changjian Li was supported by the start-up grant from the School of Informatics.

\bibliographystyle{IEEEtran}
\bibliography{main}

% \vspace{-2em}
\begin{IEEEbiography}[{\includegraphics[width=1in,height=1.25in,clip,keepaspectratio]{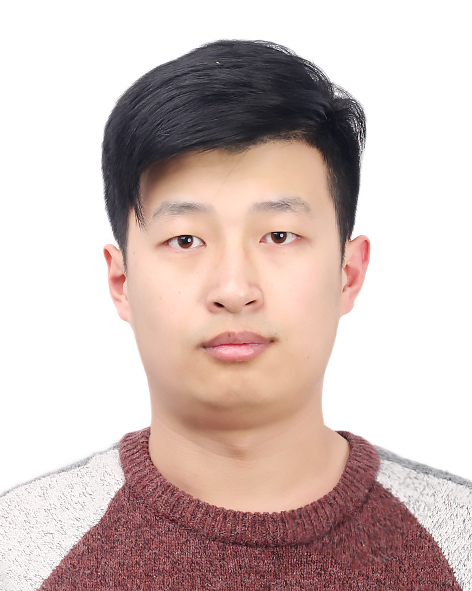}}]{Guangshun Wei} is a postdoctoral fellow at the University of Hong Kong. He received his B.Sc. degree in computer science and Technology from College of Information Science and Engineering, University of Jinan, in 2017. He received a Ph.D. from the School of Software, Shandong University, Jinan, China, in 2022. His current research interests include machine learning, image processing, and geometric analysis.
% \vspace{-1.3cm}
\end{IEEEbiography}

% \vspace{-2em}
\begin{IEEEbiography}[{\includegraphics[width=1in,clip,keepaspectratio]{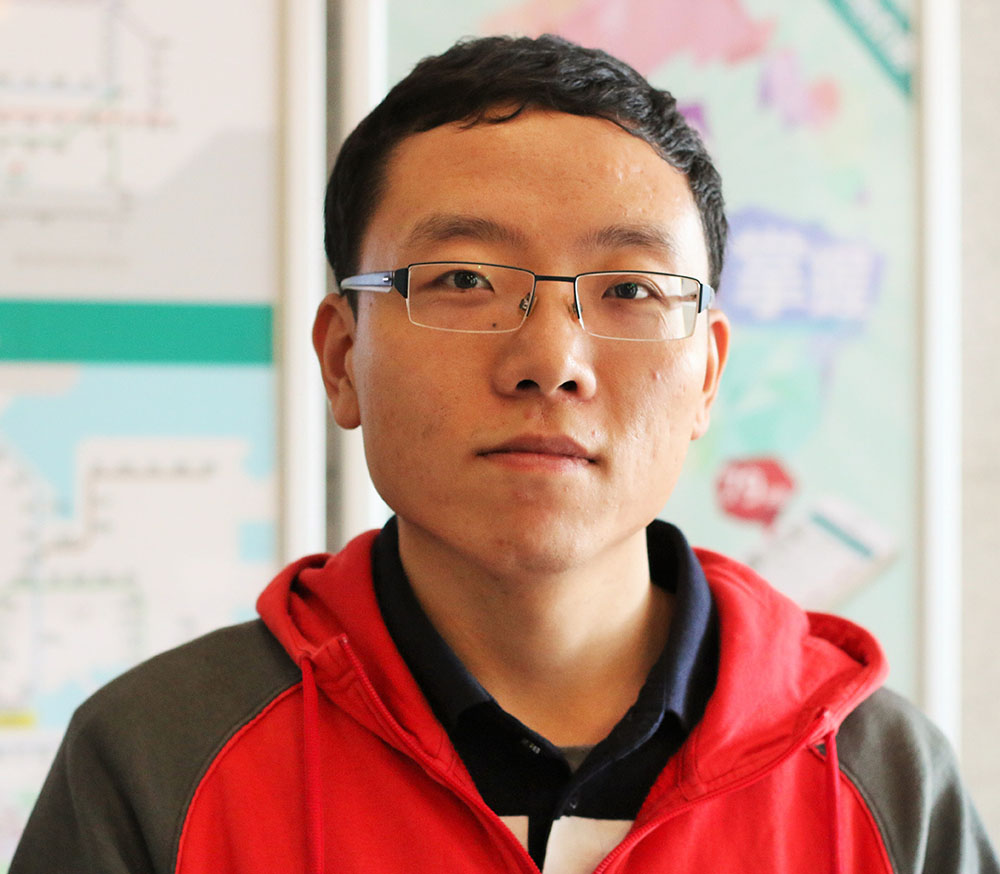}}]{Hao Pan} received the PhD degree in computer science from The University of Hong Kong. He is currently a senior researcher with Microsoft Research Asia. His research interests include computer graphics and geometric deep learning. For more information, please visit https://haopan.github.io/.
% \vspace{-1.3cm}
\end{IEEEbiography}

% \vspace{-2em}
\begin{IEEEbiography}[{\includegraphics[width=1in,height=1.25in,clip,keepaspectratio]{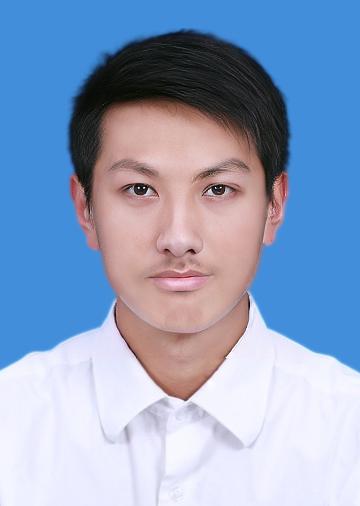}}]{Shaojie Zhuang} received a B.S. degree from the School of Software, Shandong University, in 2021. He is currently pursuing a Ph. D. degree in the School of Software, Shandong University, Jinan, China. His current research interests include machine learning, medical image processing and geometric analysis.
    \vspace{-0.8cm}
\end{IEEEbiography}

% \vspace{-2em}
\begin{IEEEbiography}[{\includegraphics[width=1in,height=1.25in,clip,keepaspectratio]{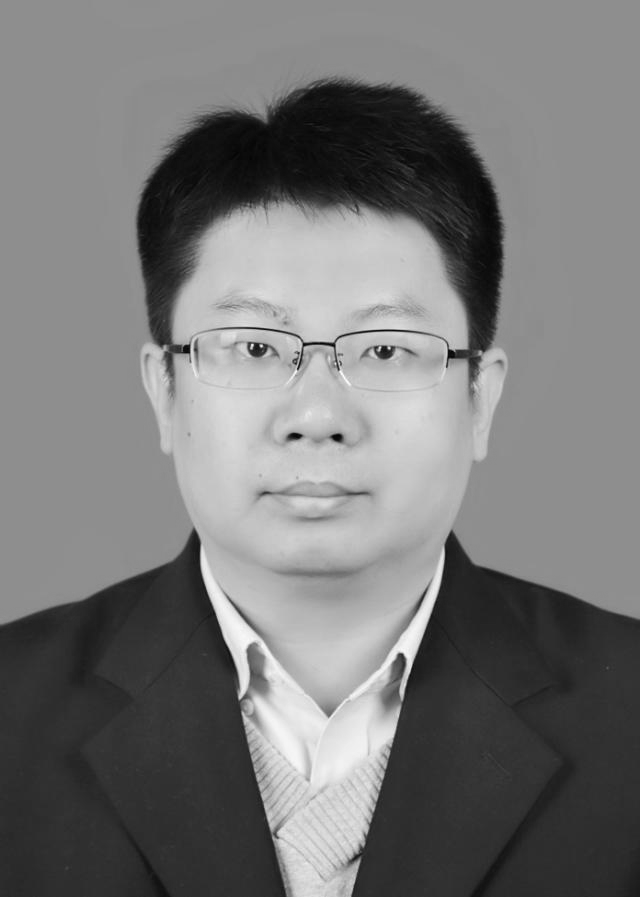}}]{Yuanfeng Zhou}
received a master’s and Ph.D. from the School of Computer Science and
Technology, Shandong University, Jinan, China,
in 2005 and 2009, respectively. He held a postdoctoral
position with the Graphics Group, Department
of Computer Science, The University
of Hong Kong, Hong Kong, from 2009 to 2011.
He is currently a Professor at the School of
Software, Shandong University, where he is also
the leader of the \textit{IGIP} Laboratory. His current
research interests include geometric modeling,
information visualization, and image processing.
\end{IEEEbiography}

\begin{IEEEbiography}[{\includegraphics[width=1in,height=1.25in,clip,keepaspectratio]{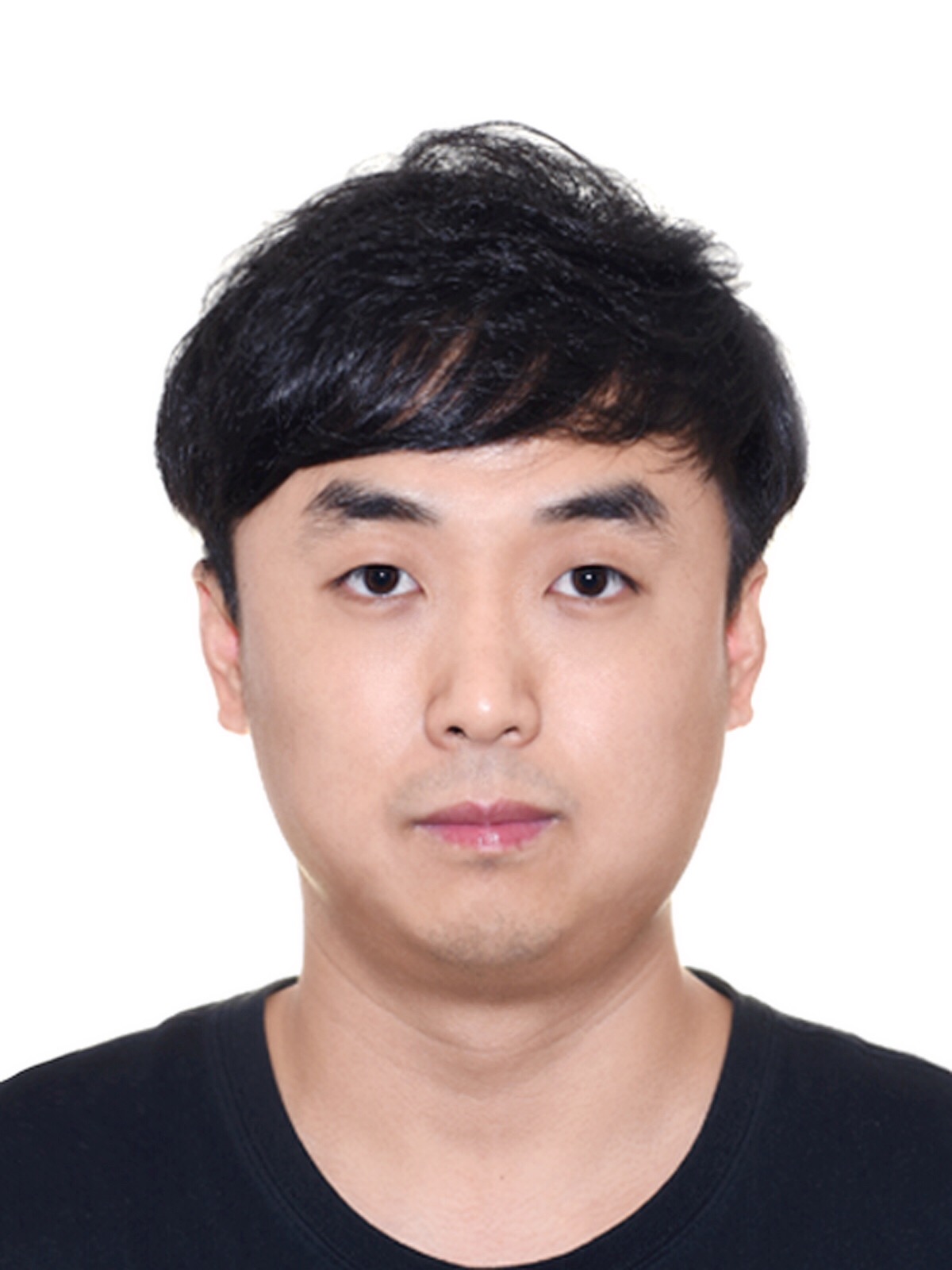}}]{Changjian Li}
received the PhD degree in Computer Science from The University of Hong Kong in 2019. He is currently an Assistant Professor of Computer Science at the University of Edinburgh. His current
research interest is mainly 3D Shape Creation and Analysis with applications in Sketch-based Modeling, Shape Reconstruction and Analysis from Point Clouds, and Medical Image Processing and Modeling. For more information, please visit https://enigma-li.github.io/.
\end{IEEEbiography}

\end{document}